%% file: main.tex
\documentclass[final]{cvpr}
\usepackage{times}
\usepackage{epsfig}
\usepackage{graphicx}
\usepackage{gensymb}
\usepackage{amsmath,amssymb} % define this before the line
\usepackage[utf8]{inputenc} % allow utf-8 input
\usepackage[T1]{fontenc}    % use 8-bit T1 fonts
\usepackage{multirow}
\usepackage[font={small}]{caption}
\usepackage{comment}
\usepackage{amsfonts}       % blackboard math symbols
\usepackage{nicefrac}       % compact symbols for 1/2, etc.
\usepackage{microtype}      % microtypography
\usepackage{url}            % simple URL typesetting
\usepackage{booktabs}       % professional-quality tables
\usepackage{amsfonts}       % blackboard math symbols
\usepackage{nicefrac}       % compact symbols for 1/2, etc.

\usepackage{xcolor}
\usepackage{subfig}

\usepackage{graphicx}
\usepackage{textcomp}
\usepackage{booktabs}
\usepackage{amsmath,amssymb,amsfonts,dsfont,pifont,bm,bbm,mathrsfs,mathtools,nicefrac}
\usepackage{algorithm,algpseudocode,listings}
\usepackage{booktabs,multirow,adjustbox,diagbox,threeparttable}

\usepackage[misc]{ifsym}

\usepackage[pagebackref=true,breaklinks=true,colorlinks,bookmarks=false]{hyperref}

\begin{document}

%%%%%%%%% TITLE
\title{AVI-Talking: Learning Audio-Visual Instructions for \\Expressive 3D Talking Face Generation}

\author{Yasheng Sun$^{1}$, Wenqing Chu$^{2}$, Hang Zhou$^{2}$, Kaisiyuan Wang$^{3}$, Hideki Koike$^{1}$ \\
    $^1$Tokyo Institute of Technology \quad $^2$Baidu Inc \quad $^3$The University of Sydney\\
      {\tt\small $\{$sun.y.aj@m, koike@c$\}$.titech.ac.jp, $\{$chuwenqing,zhouhang09,wangkaisiyuan$\}$@baidu.com}
}

\maketitle
\input{sections/abstract.tex}
\input{sections/intro.tex}

\input{sections/related.tex}
\input{sections/method.tex}

\input{sections/experiment.tex}

\input{sections/conclusion.tex}

{\small
\bibliographystyle{ieee_fullname}
\bibliography{refs}
}

\end{document}

%% file: sections/abstract.tex
\begin{abstract}
While considerable progress has been made in achieving accurate lip synchronization for 3D speech-driven talking face generation, the task of incorporating expressive facial detail synthesis aligned with the speaker's speaking status remains challenging. 
% Existing efforts either focus on learning a dynamic talking head pose synchronized with speech rhythm or aim for stylized facial movements guided by external reference such as emotional labels or reference video clips. 
% The former works often yield coarse alignment, neglecting the emotional nuances present in the audio content while the latter studies lead to unnatural applications, requiring manual style source selection by users.
Our goal is to \emph{directly leverage the inherent style information conveyed by human speech for generating an expressive talking face} that aligns with the speaking status.
In this paper, we propose \textbf{AVI-Talking}, an \textbf{A}udio-\textbf{V}isual \textbf{I}nstruction system for expressive \textbf{Talking} face generation. This system harnesses the robust contextual reasoning and hallucination capability offered by Large Language Models (LLMs) to instruct the realistic synthesis of 3D talking faces.
Instead of directly learning facial movements from human speech, our two-stage strategy involves the LLMs first comprehending audio information and generating instructions implying expressive facial details seamlessly corresponding to the speech. Subsequently, a diffusion-based generative network executes these instructions. 
This two-stage process, coupled with the incorporation of LLMs, enhances model interpretability and provides users with flexibility to comprehend instructions and specify desired operations or modifications. 
% Specifically, given a speech clip, we first employ a Q-Former for contrastive alignment the speech features with visual instructions, which is used to prompt LLMs to generate plausible visual instructions. 
% To execute these instructions, a language-guided talking face generation system with disentangled latent space is delicately derived.
% Specifically, given a speech clip, we first employ a Q-Former for contrastive alignment the speech features with visual instructions, which is then projected to input text embedding of LLMs. 
% It functions as a prompting strategy, prompting LLMs to generate plausible visual instructions that encompass diverse facial details. 
% In order to use these predicted instructions, a language-guided talking face generation system with disentangled latent space is delicately derived, where the speech content related lip movements and emotion correlated facial expressions are separately represented in \emph{speech content space} and \emph{content irrelevant space}.
% Additionally, we introduce a contrastive instruction-style alignment and diffusion technique within the content-irrelevant space to fully exploit the talking prior network for diverse instruction-following synthesis.
Extensive experiments showcase the effectiveness of our approach in producing vivid talking faces with expressive facial movements and consistent emotional status.
\end{abstract}

%% file: sections/intro.tex
\section{Introduction}
\label{sec:introduction}

%%%%%%%%%%%%%%%%%%%%%%%%%%%%%%%%%%%%%%%%%%%%%%%%%%%%%%%%%%%%5
\begin{figure}[t]
    \centering
    \includegraphics[width=0.97\linewidth]{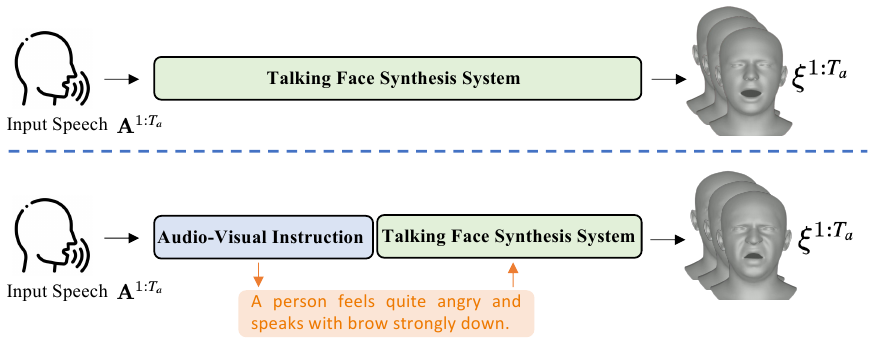}
    % \vspace{-10pt}
    \caption{ 
    In contrast to previous approaches that directly learn facial motions from speaker speech, our framework introduces an audio-visual instruction module achieved by LLMs to instruct the talking face synthesis network.
}
% \vspace{-12pt}
\label{fig:two_stage}

\end{figure}
%%%%%%%%%%%%%%%%%%%%%%%%%%%%%%%%%%%%%%%%%%%%%%%%%%%%%%%%%%%%5

Generating realistic 3D animations of human faces is crucial for a multitude of entertainment applications, encompassing digital human animation, visual dubbing in movies, and the creation of virtual avatars. 
To synthesize expressive speech-driven 3D talking face, previous work either 1) model the correlation between dynamic head poses and audio rhythm~\cite{chen2020talking,sun2023diffposetalk} or 2) borrow an external representation~\cite{ma2023styletalk,ma2023talkclip,peng2023emotalk} such as emotion labels or video clips as style reference during generation. 
However, the head dynamics hold limited expressive ability thus only yield coarse alignment, neglecting the emotional nuances present in the audio content. The latter studies require manual style source selection by users, leading to unnatural applications. 
In the paper, we explore a more natural scenario, targeting to directly leverage the underlying
style information conveyed by human speech for generating an expressive talking face that aligns with the speaking status.

Synthesizing diverse and plausible facial details based on speech while maintaining accurate lip synchronization is a highly challenging task. This challenge stems from the inherent ill-posed nature of the problem, characterized by 1) one-to-many relationship between audio inputs and potential facial movements consistent with the spoken content. Some efforts~\cite{chen2020talking,yu2023talking,ma2023dreamtalk} have introduced diffusion mechanisms to tackle diverse generation. However, direct diffusion from audio to facial motion requires bridging a huge modality gap while the information within speech and facial movements are often weakly correlated. With heavy learning burden and limited model capability, such practice is prone to capture only coarse alignment with audio cues, neglecting emotional nuances of the speaker.
2) The intertwining of the speaker's talking style and lip movements further complicates the synthesis process. Prior work~\cite{ma2023styletalk} aimed to address this entanglement by controlling specific coefficients of a parametric model. However, such practice relies on a disentangled parametric model, which is not always accessible or precise.

To handle the above challenges, we present \textbf{AVI-Talking}, an \textbf{A}udio-\textbf{V}isual \textbf{I}nstruction System for expressive \textbf{Talking} face generation.
Our key insight is to \emph{bridge the huge audio-visual modality gap with an intermediate visual instruction representation}. 
As shown in Figure.~\ref{fig:two_stage}, in contrast to previous approaches that directly learn facial movements from audio, our framework decomposes the audio-to-video generation into two stages, each with a distinctive objective, thus significantly mitigating the optimization complexities. 
Additionally, by presenting visual instruction as an intermediate output, our system enhances model interpretability and offers users with flexibility to specify their desired instructions or modifications. 

The first stage aims for comprehending the speaker talking state and imaginatively generate plausible facial expression details for subsequent instruction, necessitating robust contextual reasoning and hallucination capability.
Inspired by the impressive multi-modal understanding and generation abilities demonstrated by recent large language models (LLMs)~\cite{wu2023nextgpt,anonymous2024language}, we propose integrating LLMs as an agent~\cite{wang2023survey} to guide the talking face synthesis process.
The key aspect lies in \emph{formulating a soft prompting strategy to harness the prior contextual knowledge underlying LLMs} for speaker talking state comprehension. To achieve this, we initially employ a Q-Former to contrastively align speech features with visual instructions. Building upon the aligned audio features, we fine-tune a small number of parameters in the input projection layers for domain adaptation. Such practice not only facilitates efficient tuning but also promotes the utilization of language priors.

In the second stage, with the obtained visual instructions, our objective is to develop a speech-to-talking face network capable of synthesizing facial details that adhere to the provided instructions while preserving accurate lip movements. 
To address the inherent entanglement between lip movements and the speaker's talking style, we propose \emph{deriving a compressed latent space that distinctly identifies features related to speech content and those irrelevant to content}. By integrating both types of latent features, we can reconstruct expressive facial movements through a talking generator, thereby bypassing issues associated with inaccurate or inaccessible disentangled parametric spaces~\cite{ma2023talkclip}. 
In order to leverage this devised talking prior for instruction-following generation, it is crucial to align visual instructions within the \emph{content irrelevant} space. To facilitate joint representation learning, we introduce a contrastive instruction-style alignment and diffusion strategy. Specifically, we initially align the visual instruction contrastively to the shared content irrelevant space, upon which a diffusion prior network is employed to further refine this joint representation towards the distribution of the pre-trained talking prior.

% Instead of directly synthesizing talking face from audios, our framework follows the understand before generation paradigm. Specifically, we introduce large language model (LLMs) remarkable generative capacity to first comprehend the information implied in speech and hallucinate its described facial details. On top it, a talking face instruction system is proposed where we contrastively diffuse from instruction descriptions to latent speech-irrelevant space. 

Compared with previous studies, our main contributions in this work can be summarized as follows:
\begin{itemize}
    \item We propose AVI-Talking, an innovative audio-visual instruction system that directly leverages the inherent style information conveyed by human speech for expressive talking face generation.
    \item Large Language Models (LLMs) are introduced as an audio-visual instruction agent to comprehend speaker's talking status and generate talking face instructions.  
    \item A language-guided talking face synthesis network with disentangled speech content and content irrelevant space is designed to execute visual instruction effectively. 
    \item Experimental results validate the capability of AVI-Talking in generating vivid 3D talking faces with expressive facial details and a consistent emotional status. 
\end{itemize}

%% file: sections/related.tex
\section{Related Work}
\label{sec:related}

Speech-driven facial animation holds diverse applications in the realms of computer vision and augmented/virtual reality. This has spurred a broad spectrum of research topics encompassing both 2D~\cite{thies2020neural,wiles2018x2face,sun2021speech2talking,sun2022masked,zhou2021pose} and 3D~\cite{karras2017audio,richard2021audio,van2016conditional,fan2022faceformer,thambiraja2023imitator,xing2023codetalker,peng2023emotalk} facial synthesis. 
In the subsequent discussion, we delve into the most relevant studies in this field.

\subsection{Expressive 2D Talking Face Generation} 
Facial expressions play a pivotal role in the generation of natural talking heads~\cite{sadoughi2019speech}. 
% Diffusion-Based dreamdiffusion, difftalk, audio-visual prior, talkclip
Researchers~\cite{vougioukas2020realistic,wu2021imitating,ji2021audio,sinha2022emotion,liang2022expressive, ji2022eamm,ma2023styletalk, zhang2023sadtalker} attempt to synthesize vivid facial details while produce precise lip-synchronization.

Early approaches~\cite{danvevcek2023emotional,gan2023efficient,gururani2023space,tan2023emmn,wang2020mead} represent expressions using a limited set of emotion labels encoded as one-hot representations. To capture nuanced talking face expressions, another slew of methodologies~\cite{ji2022eamm,liang2022expressive,ma2023styletalk} incorporate reference videos as a more diverse stylistic source. While operating on RGB videos, these approaches rely on intricately designed disentanglement strategies. However, such practice often results in constrained expressiveness due to the inherent challenges of disentanglement. Meanwhile, these 2D animation stylized talking face methods have limited applicability in scenarios that demand 3D representations, such as in augmented reality (AR).

Instead of requiring users to seek out a stylized source, a more user-friendly approach involves directly exploiting speaking styles from the input audio. While some methods~\cite{ji2021audio, sinha2022emotion, xu2023high} derive networks to extract emotion labels, their capacity is limited to inferring only a discrete number of emotion classes from audio signals. Other researchers aim to achieve rhythmic synthesis by aligning head poses~\cite{chen2020talking} or expressions~\cite{yu2023talking} with audio cues. However, these efforts often result in coarse alignment without adequately considering the emotional content of the audio, leading to a lack of expressiveness. To enhance the vividness and controllability of talking head generation, recent works leverage text as an interface, allowing users to specify their desired styles~\cite{ma2023talkclip, wang2023agentavatar}. In contrast to the aforementioned approaches, we explore harnessing the generative power of large language models (LLMs) to act as a multi-modality reasoning engine. This will actively hallucinates diverse and plausible facial details based on the emotional content of the input audio, thereby offering a more comprehensive and nuanced synthesis.
% In this way, the audio-visual instruction is not explainable but also provides users flexibility to specify their desired instruction. 

\subsection{Speech-Driven 3D Talking Head Generation} 
Unlike 2D facial animation, which operates on RGB videos, 3D talking head generation employs speech-conditioned animation, utilizing geometric representations like the neural radius field (NerF)~\cite{guo2021ad} or 3D parametric templates~\cite{fan2022faceformer}.
While methods such as~\cite{fan2022faceformer, xing2023codetalker, cudeiro2019capture, peng2023emotalk} successfully synchronize facial motion with the driven audio, they often learn deterministic models, resulting in rigid motion within speech-irrelevant regions, leading to unnatural synthesis.
To address these limitations, recent approaches~\cite{sun2023diffposetalk} introduce a diffusion mechanism for its remarkable generative capability, yielding diverse high-quality synthesis results~\cite{tevet2022human, lam2022bddm}. However, while modeling various poses and expressions, these approaches neglect to capture the emotional content implied within the audio. Furthermore, methods relying on end-to-end diffusion, from reference video or style embedding to parametric models, lack explainability.
Therefore, we propose integrating a large language model into our system to firstly generate an interpretable audio-visual instruction, which is leveraged to guide the speech-driven 3D talking head generation. To augment emotion awareness, we apply the diffusion process coupled with contrastive learning solely to the speech-irrelevant space.

\subsection{LLM for Cross-Modal Learning} 
Large language models (LLMs) have demonstrated profound capabilities~\cite{wei2022chain,huang2022towards} as remarkable reasoning engines in various language generation tasks, attributed to their emergent ability~\cite{wei2022emergent}. Diverse LLMs, such as OPT~\cite{Zhang2022OPTOP}, LLaMA~\cite{Touvron2023LLaMAOA}, and GLM~\cite{zeng2023glmb}, can be fine-tuned or instructed for various purposes~\cite{NEURIPS2022_b1efde53}.
Specifically, many studies attempt to construct LLMs proficient in multi-modal reasoning and actioning~\cite{wu2023nextgpt}, leading to the emergence of MM-LLMs. 
Some studies point out that LLMs could even outperform diffusion models on standard image and video generation benchmarks~\cite{anonymous2024language}. 
In the pursuit of LLMs capable of handling both multi-modal input and output, some approaches explore employing LLMs as decision-makers~\cite{schick2023toolformer} and utilizing existing off-the-shelf multi-modal encoders and decoders as tools for executing multi-modal input~\cite{zhu2023minigpt} and output~\cite{huang2023audiogpt,xu2023secap,sun2023imagebrush}. 

Recent advancements in talking face generation have demonstrated the language model's capacity to generate multi-modal content~\cite{zheng2023minigpt5} and synthesize facial motions~\cite{ng2023can,wang2023agentavatar}. Typically, these approaches involve deriving special tokens for another modality and learning a projection layer to align them with language space~\cite{ng2023can}. However, this process demands substantial paired data and intricate training techniques for effective alignment. In contrast to these methodologies, our approach takes a direct path by predicting the text description of emotional status and facial details. This eliminates the need for challenging cross-modal alignment procedures, which also provides enhanced explainability and flexibility to users.

%% file: sections/method.tex
\section{Methodology}
\label{sec:methodology}

%%%%%%%%%%%%%%%%%%%%%%%%%%%%%%%%%%%%%%%%%%%%%%%%%%%%%%%%%%%%5
\begin{figure*}[t]
    \centering
    \includegraphics[width=1.0\linewidth]{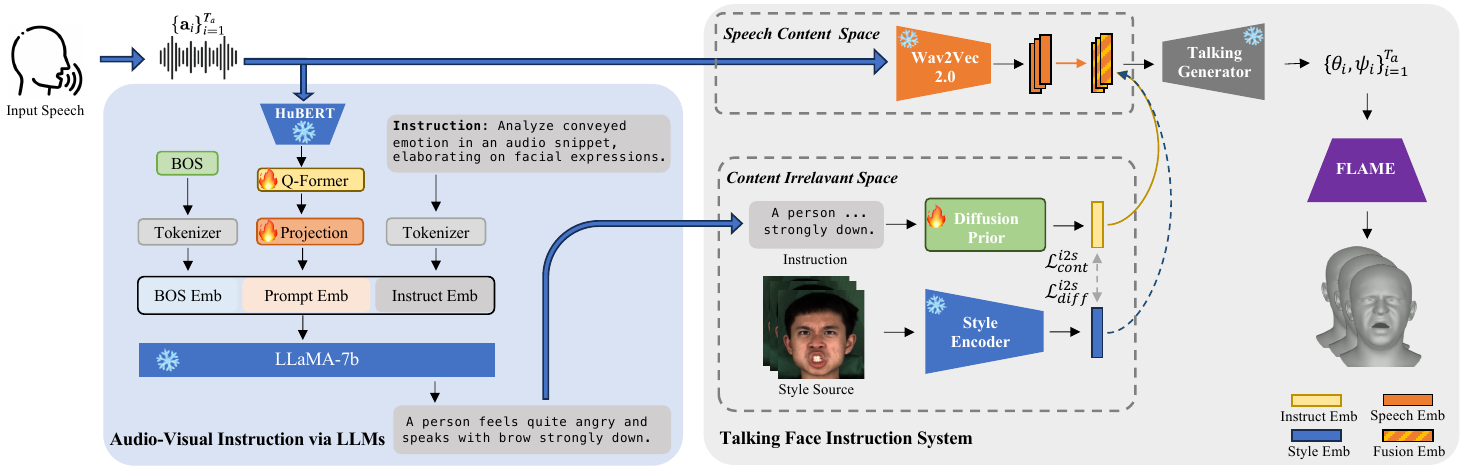}
    % \vspace{-10pt}
    \caption{The overall pipeline of our Audio-Visual Instruction Talking (AVI-Talking) Framework. Given a clip of speaker speech \{ $a_i \}_{i=1}^{T_a}$, it is first processed by Large Language Models (LLMs) to propose visual instructions encompassing plausible facial detail descriptions. Subsequently, these visual instructions, together with audio clip, are separately fed into the talking face instruction system to generate a time series of 3D parametric coefficients
    $\{\theta_i,\psi_i\}_{i=1}^{T_a}$. 
}
% \vspace{-12pt}
\label{fig:pipeline}

\end{figure*}
%%%%%%%%%%%%%%%%%%%%%%%%%%%%%%%%%%%%%%%%%%%%%%%%%%%%%%%%%%%%5

We present an~\textbf{Audio-Visual Instruction} System for Expressive~\textbf{Talking} Face Generation (\textbf{AVITalking}) that aims to achieve vivid facial expressions synthesis coherent with speaker speech status. The whole pipeline is depicted in Figure.~\ref{fig:pipeline}. In this section, we start by briefly outlining the fundamentals of the parametric 3D face model and diffusion models (Sec.~\ref{preliminary}). Subsequently, we present an overview of our pipeline (Sec.~\ref{overall_pipeline}). We then delve into the process of \emph{audio-visual instruction} utilizing Large Language Models (LLMs) (Sec.~\ref{first_stage}). Finally, we provide detailed description for \emph{instruction-following talking face synthesis} (Sec.~\ref{second_stage}). 
% As shown in Figure.~\ref{}, \textbf{AVITalking} consists of two major stages, an audio-visual instruction stage guided by large language models (LLMs) and a language-driven talking face instruction stage.
% Given a clip of input speech, it is first leveraged by LLM to analyze conveyed emotion and speculate plausible facial details, which is then fed to a talking face generation system for instruction-following synthesis.

\subsection{Preliminaries}
\label{preliminary}
\subsubsection{Parametric 3D Face Model}
Animating a template mesh that encapsulates a 3D structural representation holds promise not only for AR/VR applications but also for facilitating 2D talking face synthesis~\cite{zhang2023sadtalker}. However, the availability of 3D datasets capturing expressive facial movements is limited. Therefore, we employ a 3D reconstruction method~\cite{danvevcek2022emoca} to convert video clips from 2D audio-visual datasets~\cite{ma2023talkclip, livingstone2018ryerson} into 3D talking face datasets. Meanwhile, such practice provides both 2D images and 3D representations to enhance the training process.

Specifically, we adopt FLAME~\cite{FLAME:SiggraphAsia2017} as our template mesh. The FLAME model is a parametric 3D head model expressed as a function $\textbf{M}(\beta, \theta, \psi) \rightarrow (\textbf{V},\textbf{F})$, where the parameters consist of identity shape $\beta \in \mathbb{R}^{|\beta|}$, facial expression $\psi \in \mathbb{R}^{|\psi|}$ and pose $\theta \in \mathbb{R}^{3k+3}$ involving rotation $R \in SO(3)$ and translation $t \in \mathbb{R}^3$. After conversion, FLAME outputs a 3D mesh with vertices~$\textbf{V} \in \mathbb{R}^{n_v \times 3}$ and faces~$\textbf{F} \in \mathbb{R}^{n_f \times 3}$, where $n_v$ represents the number of vertices and $n_f$ denotes the number of faces.

\subsubsection{Diffusion Model}
% Diffusion models have become a class of popular generative models in the field of motion and image synthesis due to it diverse high-quality generation capability~\cite{tevet2022human, lam2022bddm}. The diffusion model is 
% Given $\textbf{x}_0$ from 
The goal of generative models is to learn a distribution that approximates real data distribution $q(\textbf{x}_0)$. 
The denoising diffusion probabilistic models (DDPMs)~\cite{ho2020denoising} present a multi-step progress to approximate $q(\textbf{x}_0)$ with $p_{\theta}(\textbf{x}_0)$ parameterized by $\theta$, involving both a forward and reverse process. 

The~\emph{forward process}, often referred to as~\emph{diffusion process}, transforms the real structured distribution into Gaussian noise, constructing a posterior distribution $q(\textbf{x}_{1:T}|\textbf{x}_0)$. This process follows a Markov chain that progressively introduces Gaussian noise to the data samples. Formally,
\begin{align} \label{eq:forward}
q(\bm{x}_{1:T} | \bm{x}_0) &= \prod_{t=1}^T q(\bm{x}_t | \bm{x}_{t-1}),\\
\quad q(\bm{x}_t | \bm{x}_{t-1}) &= \mathcal{N}(\bm{x}_t; \sqrt{1 - \beta_t}\bm{x}_{t-1}, \beta_t \bm{I}).
\end{align} 
Here, the constants $\beta_t$ follow an increasing trend~\cite{ho2020denoising} such that when $\beta_t$ approximate to 1, the $x_t$ approximates the Gaussian noise distribution $\mathcal{N}(0, \bm{I})$.

The~\emph{reverse process}, also known as~\emph{generative process}, targets to reverse the Gaussian noise back to joint distribution $p_{\theta}(x_{0:T})$. Formally,
\begin{align} 
\label{eq:reverse}
p_{\theta}(\bm{x}_{0:T}) &= p_{\theta}(\bm{x}_T) \prod_{t=1}^T p_{\theta}(\bm{x}_{t-1} | \bm{x}_{t}),\\
 p_{\theta}(\bm{x}_{t-1}|\bm{x}_{t}) &= \mathcal{N}(\bm{x}_{t-1}; \mu_{\theta}(\bm{x}_t, t), \Sigma_{\theta}(\bm{x}_t, t)).
\label{eq:reverse2}
\end{align} 
Here, the variance $\Sigma_{\theta}(\bm{x}_t, t) = \beta_t \bm{I}$ is set as a time-dependent constant. Therefore, a generative model $\mathcal{G}_{\theta}$ could be devised to approximate mean value of Gaussian distribution. For conditional generation, the conditional signal $\textbf{c}$ can be naturally integrated into the network architecture. Formally, the model parameters $\theta$ are optimized for all sampled timestamps $t$ and $\bm{x}$ with the following objective:
\begin{align}
    \label{eq:loss}
    \mathcal{L}_{\theta} = \mathbb{E}_{\bm{x},t}[\left\| \bm{x} - \mathcal{G}_{\theta} (\bm{x}, t, \bm{c}) \right\|^2].
\end{align}

\subsection{Overall Pipeline}
\label{overall_pipeline}
Given an audio clip, our objective is to animate a template mesh with synchronized lip movements and consistent facial expressions.
Instead of directly learning to synthesize a talking face from speech, we propose integrating Large Language Models (LLMs) to instruct the synthesis process. As illustrated in Figure.~\ref{fig:pipeline}, 
our framework, \textbf{AVI-Talking}, comprises two main stages: an audio-visual instruction stage and a talking face synthesis stage connected by visual instructions of detailed facial expression descriptions.
% \subsubsection{Problem Formulation}
Formally, our system accepts an input speech $\textbf{A}^{1:T_a}=\{ \textbf{a}_i \}_{i=1}^{T_a} $ and aims to generate a time series of 3D parametric coefficients ${\xi}^{1:T_a} = \{\theta_i,\psi_i\}_{i=1}^{T_a} $.

\subsection{Audio-Visual Instruction via LLMs}
\label{first_stage}
As illustrated on the left side of Figure~\ref{fig:pipeline}, the audio-visual instruction module takes a time series of a speaker's audio clip as input and aims to generate an instruction sentence describing detailed facial movements that conveys the individual's speaking state. The key is to \emph{develop a prompting strategy to effectively leverage the rich contextual prior knowledge inherent in LLMs}. 

Specifically, we leverage a pre-trained LLaMA as our base text generation model. In order to comprehend the speaker's speaking status existing in audio modality, the audio signal needs to be projected into text embedding of language model. 
Due to the success of pretrained-model such as HuBERT~\cite{hsu2021hubert} on Speech Emotion Recognition~\cite{mohamed2021arabic} (SER) tasks, we leverage HuBERT to encode the audio signal. Subsequently, a typical Q-Former~\cite{li2023blip,alayrac2022flamingo} architecture is employed to aggregate and extract speaking style information, bridging the gap between acoustic feature and visual facial descriptions. A linear projection layer is then learned to map the aligned feature to language model's input space. Combining the "BOS" (Beginning-of-Sequence) token with the instruction embedding, the audio prompt embedding is fed to LLaMA to prompt plausible expressive facial movements consistent with speaker status. Note that the instruction embedding is obtained by tokenizing the pre-defined instruction templates. In our experiment, we utilize instruction sentences like \emph{Analyze conveyed emotion in an audio snippet,  elaborating on facial expressions}. We manually craft 10 sentences with similar meanings and randomly sample one during the training phase.

\subsubsection{Speech Feature Compression via Learnable Queries}
%%%%%%%%%%%%%%%%%%%%%%%%%%%%%%%%%%%%%%%%%%%%%%%%%%%%%%%%%%%%5
\begin{figure}[t]
    \centering
    \includegraphics[width=1.0\linewidth]{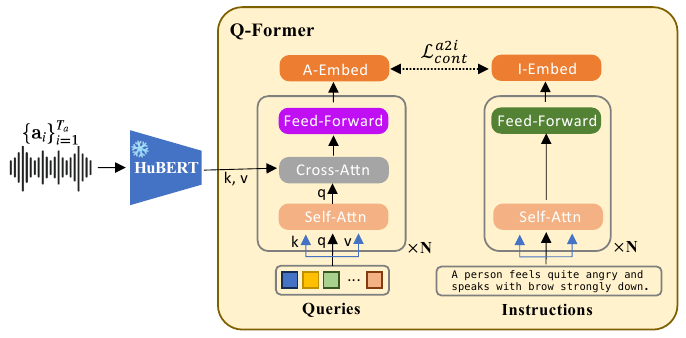}
    % \vspace{-10pt}
    \caption{The Q-Former architecture leverages the standard Perceiver network~\cite{alayrac2022flamingo} to compress speech input to a fixed-length audio embedding $\textbf{F}_{si}^{a} \in \mathcal{R}^{q_a \times l}$. 
    A contrastive loss $\mathcal{L}_{cont}^{a2i}$ is applied to encourage the queries extract audio representation that are most relevant to visual instructions.
}
% \vspace{-12pt}
\label{fig:qformer}

\end{figure}
%%%%%%%%%%%%%%%%%%%%%%%%%%%%%%%%%%%%%%%%%%%%%%%%%%%%%%%%%%%%5

The audio features extracted from HuBERT encapsulate complex information, including speech content, emotional status, and acoustic details. To effectively prompt the language model, it's essential to first comprehend and extract relevant facial movement information from the speech. Here, we employ the Q-Former architecture ~\cite{li2023blip,alayrac2022flamingo} to achieve this task.

As depicted in left side of Figure~\ref{fig:qformer}, learnable queries with fixed length are utilized to aggregate and compress speech information by cross-attention. Notably, such practice results in an audio embedding $\textbf{F}_{si}^{a} \in \mathcal{R}^{q_a \times l}$ with the same dimensionality as the query length $q_a$. This design choice simplifies the learning process and enhances generalization performance when handling speech inputs of varying lengths. Subsequently, the audio embedding is fed to a projection module for prompt embedding in language model space. To implement this, we fine-tune a small number of parameters in the input projection layers for domain adaptation.

\subsubsection{Contrastive Audio-Visual Instruction Alignment}

To eliminate unnecessary information such as speech content, environment noise and focus on extracting facial movements related feature, we adopt contrastive learning~\cite{oord2018representation} protocol to constrain the output of learned queries $\textbf{F}_{si}^{a} \in \mathcal{R}^{q_a \times l}$. The contrastive learning paradigm aligns audio embeddings and instruction features to maximize their mutual information. This is achieved by enhancing higher audio-instruction similarity of positive pairs against those of negative pairs. Specifically, we feed the corresponding instruction through a text transformer and obtain an instruction embedding as shown in the right side of Figure~\ref{fig:qformer}. Its output embedding of $\left [ CLS \right]$ token is  $\textbf{F}_{si}^{i} \in \mathcal{R}^l$. Since there are $q_a$ query embeddings, we average $\textbf{F}_{si}^{a}$ across all queries to obtain the $\bar{\textbf{F}}_{si}^a \in \mathcal{R}^l$ and apply contrastive learning as follows:
\begin{align}
    \mathcal{L}_{cont}^{a2i} = -\text{log}[\frac{\text{exp}({\mathcal{D}(\bar{\textbf{F}}_{si}^a, \textbf{F}_{si}^{i})})}{\text{exp}({\mathcal{D}(\bar{\textbf{F}}_{si}^a, \textbf{F}_{si}^{i})}) + \sum_{j=1}^{N^-}\text{exp}({\mathcal{D}(\bar{\textbf{F}}_{si}^a, \textbf{F}_{si(j)}^{i-})})}].
\end{align}
The paired in-batch samples are regarded as positive samples $(\bar{\textbf{F}}_{si}^a, \textbf{F}_{si}^{i})$ while the unpaired $N^{-}$ samples are taken as negative samples $(\bar{\textbf{F}}_{si}^a, \textbf{F}_{si(j)}^{i-})$. Here we opt for cosine distance $\mathcal{D}(\textbf{F}_1, \textbf{F}_2) = \frac{\textbf{F}_1^{\mathbf{T}} * \textbf{F}_2}{|\textbf{F}_1|\cdot|\textbf{F}_2|}$ as feature distance measurement.

\subsubsection{Instruction Generation via Projection Layer Finetuning}
After the Q-Former is pre-trained to contrastively align acoustic features to visual facial descriptions. Subsequently, the Q-Former is frozen, and we fine-tune the input linear projection layer of LLaMA-7b to achieve visual instruction prediction as shown in Figure~\ref{fig:pipeline}. 
Specifically, We follow the general text generation training paradigm~\cite{zhu2023minigpt} to learn this projection layer.

\subsection{Instruction-Following Talking Face Synthesis}

%%%%%%%%%%%%%%%%%%%%%%%%%%%%%%%%%%%%%%%%%%%%%%%%%%%%%%%%%%%%5
\begin{figure}[t]
    \centering
    \includegraphics[width=1.0\linewidth]{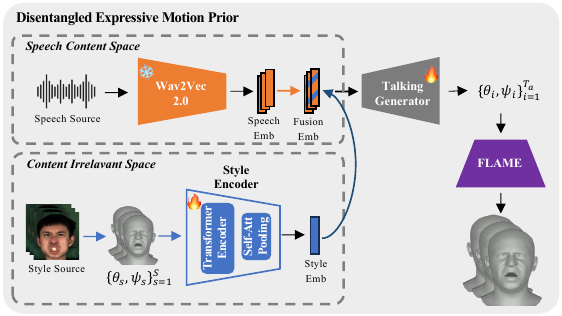}
    % \vspace{-10pt}
    \caption{
    To establish a disentangled expressive motion prior, we learn two complementary latent spaces, \emph{speech content} space and \emph{content irrelevant} space. In \emph{speech content} space, we represent lip movements related to speech content, while in the \emph{content irrelevant} space, we capture facial expressions correlated with the speaking state. 
}
\vspace{-12pt}
\label{fig:motion_prior}

\end{figure}
%%%%%%%%%%%%%%%%%%%%%%%%%%%%%%%%%%%%%%%%%%%%%%%%%%%%%%%%%%%%5

\label{second_stage}
With the obtained facial instructions, a talking face synthesis network aims to animate a mesh template with synchronized lip movements and expressions as illustrated on the right side of Figure~\ref{fig:pipeline}. The movements of the lips and facial expressions exhibit a high degree of correlation with each other~\cite{ma2023styletalk}. For example, specific pronunciations often convey relevant emotions.
To address this correlation and potential entanglement, we propose initially training a disentangled talking prior~\cite{richard2021meshtalk,danvevcek2023emotional}, wherein the speech content space and content irrelevant space are distinguished (shown in Figure~\ref{fig:motion_prior}). 
Subsequently, a diffusion prior module (shown in Figure~\ref{fig:contrastive_diffusion}) is devised to bridge the gap between instruction text and talking styles within the identified content irrelevant space. 
% The outputs of the diffusion prior are mapped to the content irrelevant space, which can be utilized by a pre-trained talking generator for synthesizing facial movements.

\subsubsection{Disentangled Expressive Motion Prior}

As depicted in Figure.~\ref{fig:motion_prior}, we target to establish a disentangled latent space, where the speech content related lip-movements and facial expressions correlated with speaking state are distinctly represented in \emph{speech content} space and \emph{content irrelevant} space, respectively.
Concretely, in speech content space we employ a pretrained ASR network, Wav2Vec 2.0~\cite{baevski2020wav2vec} to encode the speaker audio $\textbf{A}^{1:T_a}$. 
These extracted speech features capture semantic content information, which is subsequently utilized by the talking generator for syllable pronunciation.
In order to encode additional talking style information, we point out the existence of \emph{content irrelevant} space for representing content-repelling information such as talking styles, poses and speaker identity. 

To learn the \emph{content irrelevant} space, we employ a transformer-based style encoder~\cite{ma2023styletalk} designed to capture content-repelling information. For a given talking video, we randomly select $S$ reference frames to serve as the source for the speaking state. These frames are then processed by the FLAME model to obtain coefficients $\{\theta_s,\psi_s\}_{s=1}^{S}$, where the coefficient at time $t$ is excluded. Subsequently, these coefficients are fed into the style encoder to extract a comprehensive speaking state representation for the video. To successfully predict coefficients at the current time step $\{\theta_t,\psi_t\}$, we rely on both the speech feature $\textbf{A}^t$ in the \emph{speech content} space and the extracted style information in the \emph{content irrelevant} space. The complementary nature of these properties naturally facilitates the learning of disentangled spaces.

\subsubsection{Contrastive Instruction-Style Alignment and Diffusion}
%%%%%%%%%%%%%%%%%%%%%%%%%%%%%%%%%%%%%%%%%%%%%%%%%%%%%%%%%%%%5
\begin{figure}[t]
    \centering
    \includegraphics[width=1.0\linewidth]{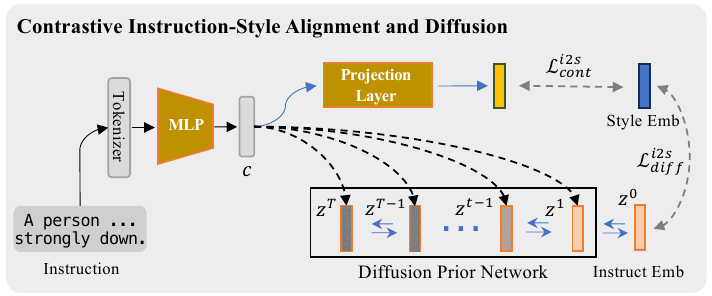}
    % \vspace{-10pt}
    \caption{Within the content irrelevant space, we contrastively align the visual instruction with style embedding to obtain a aligned feature $c$,
    upon which a diffusion prior network is employed to further
    map it towards the distribution of the
    pre-trained talking prior.
}
% \vspace{-12pt}
\label{fig:contrastive_diffusion}

\end{figure}
%%%%%%%%%%%%%%%%%%%%%%%%%%%%%%%%%%%%%%%%%%%%%%%%%%%%%%%%%%%%5

Once the content irrelevant space is identified, a natural way for cross-modality generation is to map visual instruction to the representation within this space~\cite{ma2023talkclip}.
% The motion diffusion prior network is derived to transform instruction text to talking style space. In contrast to directly accomplish instruction text-to-facial movement generation, we train this text-to-style generation process following DALL-E2~\cite{ramesh2022hierarchical} learning paradigm. 
As depicted in Figure~\ref{fig:contrastive_diffusion}, a Multi-Layer-Perceptron (MLP) network is derived to first align latent instruction representation with style embedding.
The typical contrastive loss $\mathcal{L}_{cont}^{i2s}$ is employed, following standard CLIP training process~\cite{ramesh2022hierarchical}, which we omit here.
However, since this multi-modal contrastive learning strategy only pushes the instruction embeddings to hold close direction with their associated style image features, which is prone to cause disjoint embeddings due to the existance of modality gap~\cite{liang2022mind}. 
To further activate motion prior that expects visual style embeddings, we introduce a diffusion prior network to bridge the modality gap by mapping to their distributions.

For the diffusion prior network $\mathcal{F}_{\theta}$, we leverage the typical decoder-only Transformer architecture to iteratively predict the denoised style embedding $\textbf{z}^t$ conditioned on the above representation $\textbf{c}$. 
Instead of imposing error prediction formulation~\cite{ho2020denoising}, we directly train the network to predict unnoised style embedding $\textbf{z}$ from noised embedding $\textbf{z}^t$ sampled at time step $t$. Formally,
\begin{align}
    \mathcal{L}_{diff}^{i2s} = \mathbb{E}_{\bm{\textbf{z}},t}[\left\| \bm{\textbf{z}} - \mathcal{F}_{\theta} (\bm{\textbf{z}}, t, \bm{\textbf{c}}) \right\|^2]
\end{align}
where we apply the naive Mean-Square Error (\textbf{MSE}) to the prediction result.

Therefore, the overall learning objective of visual instructions to speaking styles generation can be written as
\begin{align}
    \mathcal{L}^{i2s} = \mathcal{L}_{cont}^{i2s} + \lambda^{i2s} \mathcal{L}_{diff}^{i2s},
\end{align}
where $\lambda$ is balancing coefficients.

%% file: sections/experiment.tex
\section{Experiments}
\label{sec:experiments}

% Table 1%%%%%%%%%%%%%%%%%%%%%%%%%%%%%%%%%%%%%%%%%%%%%%%%%
\setlength{\tabcolsep}{14pt}
\begin{table*}[t] %\footnotesize
\begin{center}  \caption{The quantitative results on MeadText~\cite{ma2023talkclip} and RAVEDESS~\cite{livingstone2018ryerson}. For all approaches, we compare them under three metrics including FID~\cite{heusel2017gans}, KID~\cite{ren2023diffusion} and LSE-D~\cite{prajwal2020lip}. Lower scores indicate better performance.}

\label{table:main_exp}
\begin{tabular}{lcccccccccc}
\toprule & \multicolumn{3}{c}{MeadText~\cite{ma2023talkclip}}& \multicolumn{3}{c}{RAVEDESS~\cite{livingstone2018ryerson}} \\
\cmidrule(lr){2-4} \cmidrule(lr){5-7}
Method & $\text{FID}  \downarrow $& $\text{KID}  \downarrow $ & LSE-D $\downarrow$ & $\text{FID}  \downarrow $& $\text{KID}  \downarrow $ & LSE-D $\downarrow$ \\

\midrule
MeshTalk~\cite{richard2021meshtalk}  &201.06 &0.3601 &10.51 &134.47 & 0.2831 & 9.19 \\ % 0.0069
EmoTalk~\cite{peng2023emotalk}       &124.41 &0.2118 &\textbf{8.37} &122.95 & 0.1929 & \textbf{8.51} \\ % 0.0067
CodeTalker~\cite{xing2023codetalker} &68.68  &0.0658 &\underline{8.38} &\underline{46.90} & \underline{0.0711} & 8.99 \\
FaceFormer~\cite{fan2022faceformer}  &\underline{68.35}  &\underline{0.0611} &9.08 &47.78 & 0.0721 & 8.85 \\
GT  & -  & - & 9.36 & - & - & 9.05 \\
\hline
\textbf{AVI-Talking}  & \textbf{12.53} & \textbf{0.0190} &9.06 & \textbf{15.94} & \textbf{0.0225} & \underline{8.81} \\ % 0.0029
\bottomrule
\end{tabular}
\end{center}

\end{table*}
%%%%%%%%%%%%%%%%%%%%%%%%%%%%%%%%%%%%%%%%%%%%%%%%%%%%%%%%%%

%%%%%%%%%%%%%%%%%%%%%%%%%%%%%%%%%%%%%%%%%%%%%%%%%%%%%%%%%%%%5
\begin{figure*}[t]
    \centering
    \includegraphics[width=1.0\linewidth]{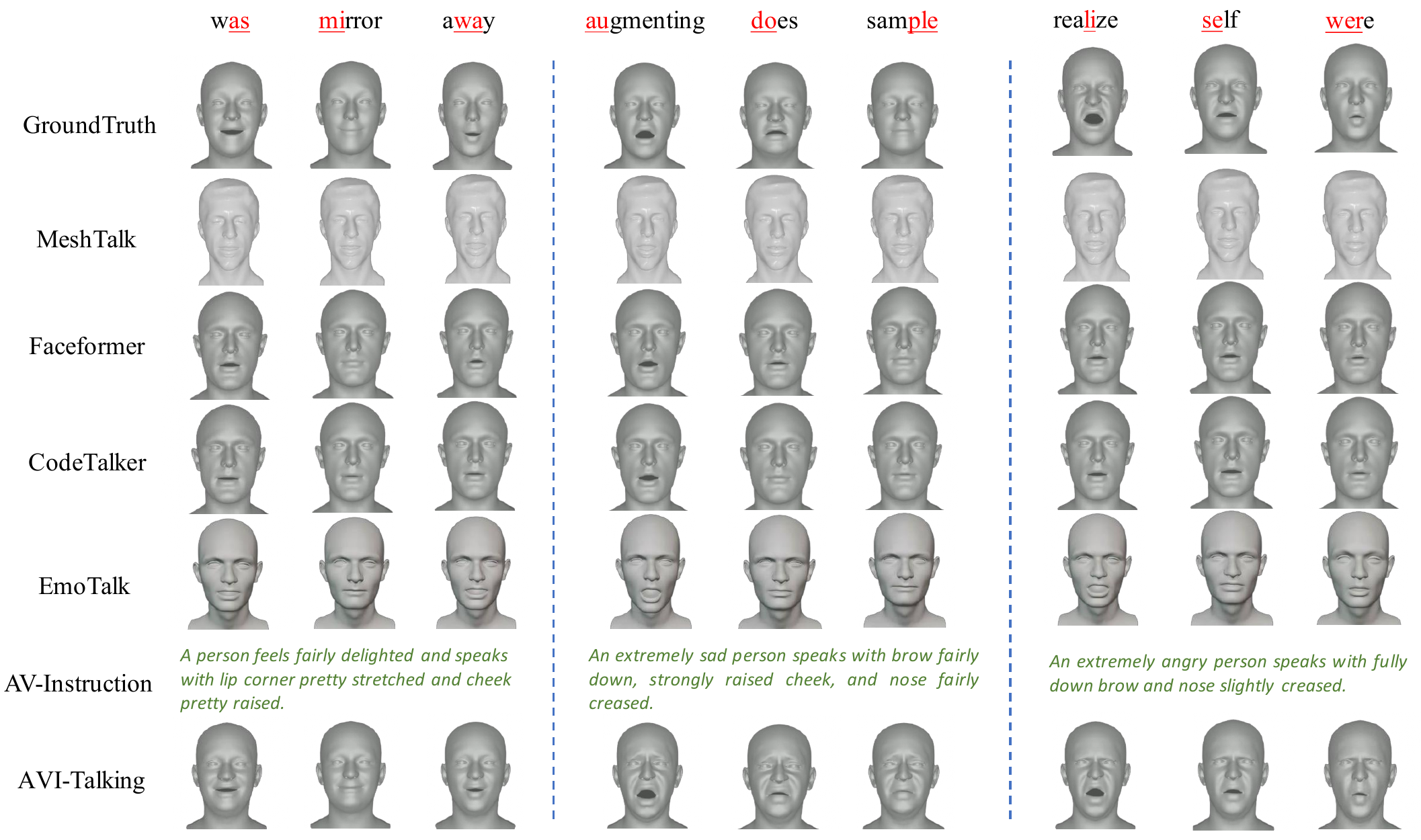}
    % \vspace{-10pt}
    \caption{Qualitative Results. In the top row are ground truth videos.
    Our generated audio-visual instructions are shown in green line. 
    In the bottom row demonstrates synthesis results guided by above instructions.
    % It can be seen that our framework create precise lip synchronization results. 
    Compared to other competitive approaches, our method achieves superior detailed expressions. Notably, our system is capable of generate facial movements distinct from Ground Truth but convey consistent speaking state (See second and third case). 
}
% \vspace{-12pt}
\label{fig:quali}

\end{figure*}
%%%%%%%%%%%%%%%%%%%%%%%%%%%%%%%%%%%%%%%%%%%%%%%%%%%%%%%%%%%%5

%%%%%%%%%%%%%%%%%%%%%%%%%%%%%%%%%%%%%%%%%%%%%%%%%%%%%%%%%%%%%%%%%%%%%%
\setlength{\tabcolsep}{8pt}
\begin{table*}[t] %\normalsize 
\begin{center}

\caption{User study measured by Mean Opinion Scores. Larger is higher, with the maximum value to be 5.}
% \vspace{-10pt}
\label{table:MOS}
\begin{tabular}{ccccc}

%\toprule
\hline
%\noalign{\smallskip}

MOS on $\setminus$ Approach & MeshTalk~\cite{richard2021meshtalk} & EmoTalk~\cite{peng2023emotalk} & CodeTalker~\cite{xing2023codetalker} & \textbf{AVI-Talking (Ours)} \\
\noalign{\smallskip}
\hline
%\midrule

Lip Sync Quality & 2.43 & 2.83 & 3.13 & \textbf{3.23} \\
Movement Expressiveness& 2.83 & 3.0 & 2.53 & \textbf{3.27} \\
Expression Consistency& 2.37 & 3.03 & 2.33 & \textbf{3.50} \\
\hline
\end{tabular}
\end{center}
\vspace{-15pt}
\end{table*}
\setlength{\tabcolsep}{1.4pt}
%%%%%%%%%%%%%%%%%%%%%%%%%%%%%%%%%%%%%%%%%%%%%%%%%%%%%%%%%%%%%

\subsection{Experimental Settings}

\subsubsection{Datasets}

We train both audio-visual instruction module and talking face instruction network on MeadText~\cite{ma2023talkclip} dataset. Evaluation is conducted on test set of RAVEDESS~\cite{livingstone2018ryerson} and MeadText. Since both datasets are made of RGB videos, we obtain reconstruction results by Emoca~\cite{danvevcek2022emoca} and render the facial meshes as GT videos for comparison.

\begin{itemize}
    \item \textbf{MeadText}~\cite{ma2023talkclip}. This dataset is extended from Mead~\cite{kaisiyuan2020mead} dataset by labeling the speaker emotional status and facial action unit (FAU) with natural language descriptions. MEAD~\cite{kaisiyuan2020mead} is a high-quality emotional talking-face dataset, including recorded videos of different actors speaking with 8 different emotions at 3 intensity levels.
    
    \item \textbf{RAVEDESS}~\cite{livingstone2018ryerson}. There are a total of 24 professional actors (12 female, 12 male) covering over 1440 utterances in a neutral North American accent. 8 speech emotions includes calm, happy, sad, angry, fearful, surprise, disgust and neutral expressions are produced at two levels of emotional intensity (normal, strong). For convenience, we choose speech videos of the first 6 actors as the evaluation dataset.

\end{itemize}

\subsubsection{Implementation Details}
The videos are sampled at a rate of 25 FPS, and the audios are pre-processed to 16 kHz for all stages of our system. The training of the audio-visual instruction module is divided into two stages. In the first stage, the audios are fed to HuBERT~\cite{hsu2021hubert} for speech feature extraction. Then, the Q-Former is pre-trained to contrastively align acoustic features to visual facial descriptions. Subsequently, the Q-Former is frozen, and we fine-tune the input projection layer of LLaMA-7b to achieve caption prediction. To enhance model performance, we leverage common text data augmentation techniques such as synonym replacement during the training stage.

For the talking face synthesis network, we adopt the model architecture of EMOTE~\cite{danvevcek2023emotional} as our basic facial motion generation network. We adapt the framework with disentangled speech content space and content irrelevant space. For speech content extraction, we utilize the state-of-the-art pretrained ASR network Wav2Vec 2.0~\cite{baevski2020wav2vec} to extract the raw waveform and compress features with temporal convolutions following a similar protocol to EMOTE~\cite{danvevcek2023emotional}. 
For speech style extraction, we follow the architecture design of StyleTalk~\cite{ma2023styletalk} and leverage the linear styling network from EMOTE~\cite{danvevcek2023emotional} as a teacher network for knowledge distillation.
Within the content irrelevant space, the training schedule of our contrastive instruction-style alignment and diffusion module is adapted from DALL-E2~\cite{ramesh2022hierarchical}'s open-source implementation of diffusion prior. Specifically, the diffusion loss weight $\lambda^{i2s}$ is set to 30 to balance optimization loss. Similar to the first stage, we also employ the same data augmentation approach to facilitate robust performance. As our focus in this work is on modeling speaking styles, the poses and speaker identity are set to a neutral state during both the training and inference stages. Both our models are implemented in PyTorch~\cite{paszke2019pytorch} and trained using 80G Tesla A100 GPUs.

\subsubsection{Comparison Methods}
We compare our methods with state-of-the-art template-based models that support speech conditional 3D talking face generation, including MeshTalk~\cite{richard2021meshtalk}, FaceFormer~\cite{fan2022faceformer}, CodeTalker~\cite{xing2023codetalker}, and EmoTalk~\cite{peng2023emotalk}.

MeshTalk~\cite{richard2021meshtalk} introduces a cross-modality disentanglement mechanism to generate realistic face animation.
FaceFormer~\cite{fan2022faceformer} devises a transformer-based architecture capable of synthesizing realistic 3D facial motions.
CodeTalker\cite{xing2023codetalker} incorporates the codebook technique\cite{esser2020taming} to enhance the accuracy of lip movements.
EmoTalk~\cite{peng2023emotalk} employs an emotional disentanglement strategy using one-hot emotional labels for face animation.

% Table 3%%%%%%%%%%%%%%%%%%%%%%%%%%%%%%%%%%%%%%%%%%%%%%%%%%%%%%%%%%%%
\setlength{\tabcolsep}{2pt}
\begin{table}[t] %\footnotesize
\begin{center}  \caption{Ablation over model design of Audio-Visual Instruction stage.}

\label{table:ablation_caption}
\begin{tabular}{lcccccccccc}
\toprule
% \cmidrule(lr){1-3}
Metric & w/o Aug. & w/o LLaMA & w/o Q-Former & Full \\

\midrule
$BLEU_{1} \uparrow $ & 45.4 & 32.4 & 41.4 & \textbf{47.4} \\
$BLEU_{4} \uparrow$ & 12.7 & 7.1 & 10.4 & \textbf{11.4} \\
$METEOR \uparrow $ & 21.5 & 16.1 & 20.7 & \textbf{22.0} \\
$ROUGE_{l} \uparrow $ & 38.0 & 28.0 & 36.0 & \textbf{38.4} \\
$CIDEr \uparrow $ & \textbf{54.5} & 32.8 & 53.0 & 49.3 \\
$SPICE \uparrow $ & 34.8 & 27.4 & 36.4 & \textbf{40.9} \\
\bottomrule
\end{tabular}
\end{center}

\end{table}
%%%%%%%%%%%%%%%%%%%%%%%%%%%%%%%%%%%%%%%%%%%%%%%%%%%%%%%%%%

% Table 4%%%%%%%%%%%%%%%%%%%%%%%%%%%%%%%%%%%%%%%%%%%%%%%%%%%%%%%%%%%%
\setlength{\tabcolsep}{4pt}
\begin{table}[t] %\footnotesize
\begin{center}  \caption{Ablation over model design of Talking Face Synthesis stage.}

\label{table:ablation_talking}
\begin{tabular}{lcccccccccc}
\toprule
% \cmidrule(lr){1-3}
Method & LSE-D $\downarrow$ & Diversity $\uparrow$ & FID $\downarrow$ & KID $\downarrow$ \\

\midrule
w/o Aug. &  9.11 & 0.433 & 14.18 & 0.0192 \\
w/o Diffusion & 9.21 & 0 & 18.72 & 0.0268 \\
w/o Cont Align & 9.07 & 0.373 & 13.37 & 0.0190 \\
Full (Ours)  &\textbf{9.06} & \textbf{0.435} & \textbf{12.53} & \textbf{0.0190} \\
\bottomrule
\end{tabular}
\end{center}

\end{table}
%%%%%%%%%%%%%%%%%%%%%%%%%%%%%%%%%%%%%%%%%%%%%%%%%%%%%%%%%%

\subsection{Quantitative Evaluation}
\subsubsection{Evaluation Metric}
We validate our method from the perspectives of both instruction generation capability and talking face synthesis quality.

\begin{itemize}
    \item \textbf{Audio-Visual Instruction Prediction.} Metrics that have popularly been involved in the field of natural language generation (NLG) task are chosen to evaluate our method. Specifically, we include $BLEU_1$, $BLEU_4$~\cite{papineni2002bleu}, $METEOR$~\cite{banerjee2005meteor}, $ROUGE_l$~\cite{lin2004rouge}, $CIDEr$~\cite{wang2019describing} and $SPICE$~\cite{anderson2016spice}.
    
    \item \textbf{3D Talking Face Synthesis.} To assess visual fidelity, we utilize standard GAN metrics:~\textbf{FID}~\cite{heusel2017gans} and~\textbf{KID}~\cite{ren2023diffusion} on face regions of rendered images. Additionally, to evaluate generation diversity, we report~\textbf{Diversity} scores~\cite{aneja2023facetalk}, measuring the extent of expression diversity generated for a given clip of human speech. Specifically, distances across predicted style features for the same audio with different noises are calculated. Moreover, we adopt~\textbf{LSE-D}~\cite{prajwal2020lip} to evaluate lip synchronization performance.

\end{itemize}

\subsubsection{Evaluation Results}
Regarding the synthesis of talking faces, our study reports quantitative results for MeadText~\cite{ma2023talkclip} and RAVEDESS~\cite{livingstone2018ryerson} in Table~\ref{table:main_exp}. Notably, our method demonstrates outstanding performance across most metrics on both datasets. However, our approach may exhibit comparatively weaker lip-sync performance, particularly in terms of LSE-D, when compared to other methods. We attribute this discrepancy partly to the strong preference bias for neutral expressions in SyncNet~\cite{prajwal2020lip}, which is pre-trained on predominantly expressionless videos. Unlike these methods, our synthesis results encompass expressive facial details, potentially leading to lower scores. Furthermore, our approach achieves LSE-D scores close to those of ground truth videos on both datasets, suggesting robust generation of precise lip-sync videos.

\subsection{Qualitative Evaluation}

\subsubsection{Qualitative Analysis}
Subjective evaluation is crucial for validating model performance in generative tasks. We encourage readers to refer to our supplementary materials for additional demo videos and comparison results. In Figure~\ref{fig:quali}, we present comparison results of our method against previous state-of-the-art approaches in three cases. It can be seen that our method produces plausible audio-visual instructions and generates expressive facial details aligned with the speaker's state.
Regarding lip synchronization performance, we observe that CodeTalker~\cite{xing2023codetalker} or Faceformer~\cite{fan2022faceformer} may generate more natural pronunciation in expressionless states. However, when involving emotional states, slight distortions in lip movements can be observed (e.g., the stretching of lip corners during happy emotions). This observation aligns with the LSE-D scores in the quantitative evaluation presented in Table~\ref{table:main_exp}. Nevertheless, our approach still achieves competitive synthesis results compared to others and approaches the performance of ground truth videos, thus validating the effectiveness of our approach in lip synchronization.

\subsubsection{User Study}
We conducted a user study involving 15 participants to gather their opinions on 30 videos generated by our method alongside three competing methods. Among these, twenty videos were created using randomly selected speaker audios from the test set of MeadText, while the remaining ten were sourced from RAVEDESS. We utilized the well-established Mean Opinion Scores (MOS) rating protocol. Participants were tasked with providing ratings on a scale of 1 to 5 for three specific aspects of each video: (1) Lip Sync Quality, (2) Movement Expressiveness, and (3) Expression Consistency. Lip sync quality evaluates mouth movements in sync with speech content, movement expressiveness assesses facial detail richness, and expression consistency measures the alignment between facial movements and speaker speech expressions.

The results are presented in Table~\ref{table:MOS}. MeshTalk~\cite{richard2021meshtalk} scores the lowest across all aspects, possibly attributed to the architecture design of its naive UNet. By incorporating transformer blocks, EmoTalk~\cite{peng2023emotalk} and CodeTalker~\cite{xing2023codetalker} achieve higher lip-sync scores. Regarding movement expressiveness and expression consistency, our model significantly surpasses other approaches, owing to its carefully derived audio-visual instruction strategy. Overall, our AVI-Talking model outperforms its counterparts in expressive synthesis, highlighting the effectiveness of our approach.

\subsection{Further Analysis}
%%%%%%%%%%%%%%%%%%%%%%%%%%%%%%%%%%%%%%%%%%%%%%%%%%%%%%%%%%%%5
\begin{figure}[t]
    \centering
    \includegraphics[width=1.0\linewidth]{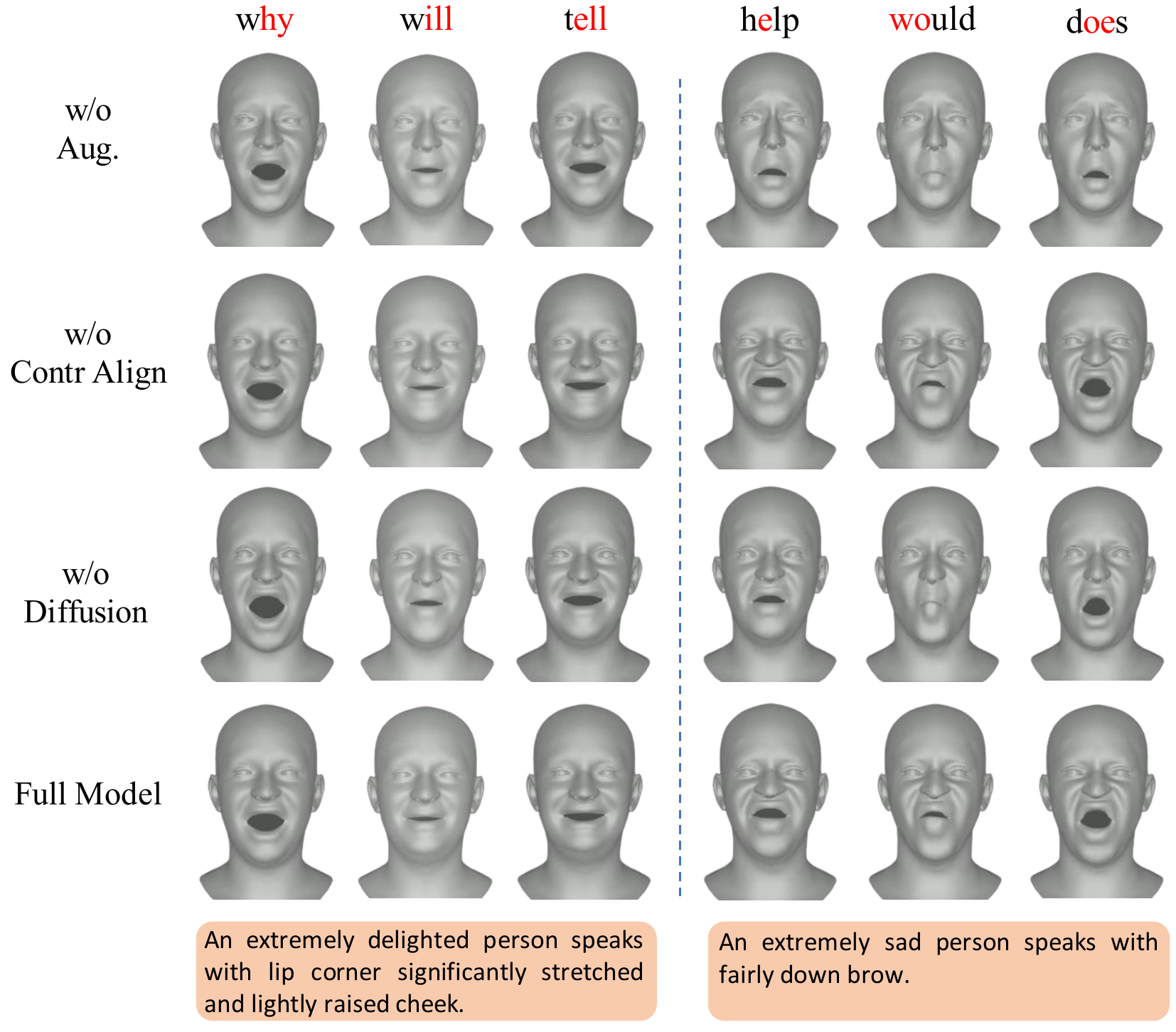}
    % \vspace{-10pt}
    \caption{Ablation Study. The bottom row illustrates the audio-visual instruction, while the rows above visualize the generation results across three key aspects of model design. Without diffusion, the model tends to produce conservative results, thereby inadequately raising the lip corner during smiling. Not utilizing data augmentation can result in sub-optimal convergence, failing to capture precise detailed facial movements.
}
\vspace{-8pt}
\label{fig:ablation}

\end{figure}
%%%%%%%%%%%%%%%%%%%%%%%%%%%%%%%%%%%%%%%%%%%%%%%%%%%%%%%%%%%%5
\subsubsection{Ablation Study}
We conduct ablation studies on both stages of our system, wherein we systematically remove three crucial components from each stage to evaluate the effectiveness of our framework design.

\textbf{Audio-Visual Instruction Module.} 
We conduct experiments on the first stage model (1) w/o text augmentation; (2) w/o LLaMA generator and (3) w/o Q-Former alignment. 
For the setting without the LLaMA base model, we adopt the BLIP2 training paradigm \cite{li2023blip} and utilize image-grounded text generation loss for instruction generation. The numerical results on the MeadText dataset \cite{ma2023talkclip} are presented in Table \ref{table:ablation_caption}. We find that without text data augmentation, the model tends to overfit to a sub-optimal point, leading to slightly worse performance. Removing the LLaMA model results in the loss of rich contextual knowledge, thereby also causing inferior performance. Furthermore, without the Q-Former contrastive alignment strategy, the extraction and alignment of speech features to text embedding become inadequate, introducing significant training difficulties and resulting in significantly inferior performance.

\textbf{3D Talking Face Synthesis.} 
For the second stage, we train and evaluate the talking face synthesis network by removing (1) text augmentation, (2) the diffusion prior network, and (3) contrastive alignment. The numerical results on the MeadText dataset \cite{ma2023talkclip} are demonstrated in Table \ref{table:ablation_talking}, and visualization results are depicted in Figure \ref{fig:ablation}. Similar to the first stage, without text data augmentation, the synthesis results suffer from inferior performance on all metrics. Visualization results in the first row illustrate that the absence of augmentation tends to inadequately capture the smiling lip corner motion (See the first case in the left column). Without employing the diffusion strategy, the generation process becomes deterministic, leading to a lack of diversity. We also observe significantly reduced performance on other metrics, possibly due to the diverse generation nature of this problem. Visualizations in Figure \ref{fig:ablation} indicate that without adopting the diffusion strategy, the network tends to produce conservative generations, where the lip corner is not as well stretched as in our full model (See the first case in the left column). Removing contrastive alignment also results in inferior outcomes, highlighting its effectiveness in boosting generation performance.

\subsubsection{Visualization of Aligned Speech Features}
%%%%%%%%%%%%%%%%%%%%%%%%%%%%%%%%%%%%%%%%%%%%%%%%%%%%%%%%%%%%5
\begin{figure}[t]
    \centering
    \vspace{-20pt}
    \includegraphics[width=1.0\linewidth]{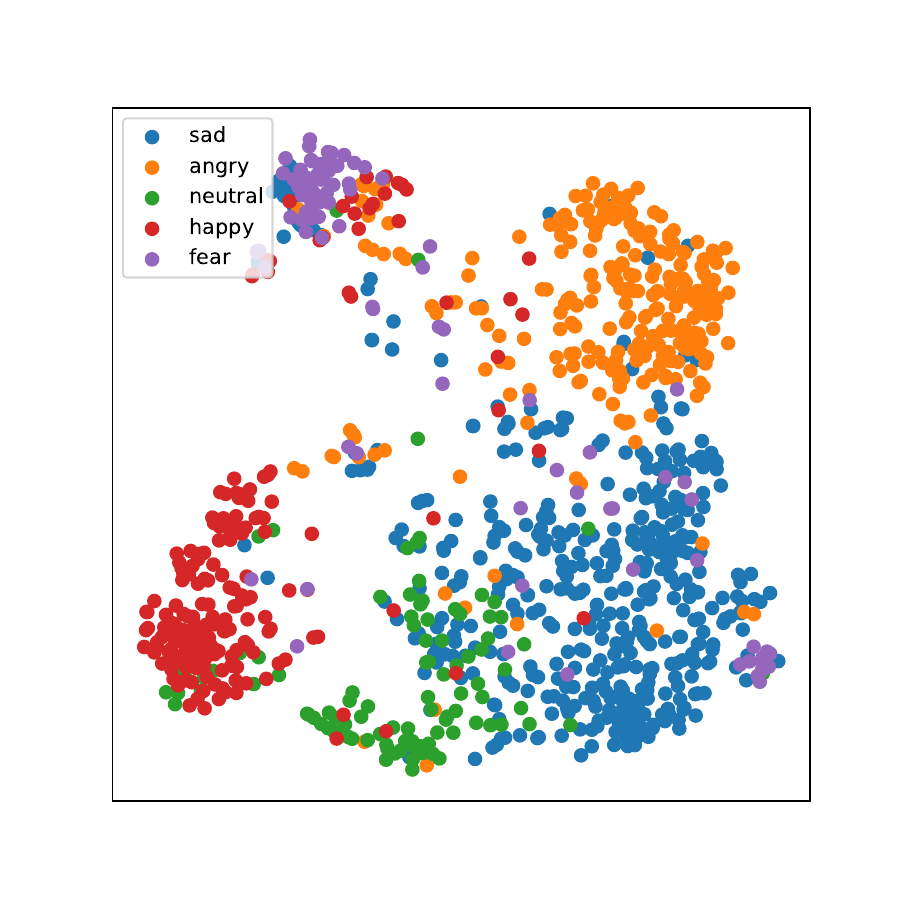}
    \vspace{-35pt}
    \caption{Visualizations of t-SNE embeddings derived from aligned speech features using Q-Former. The audio samples are from male utterances in the MeadText dataset, focusing on five typical speaking emotions. Notably, the aligned speech features corresponding to each specific emotion exhibit closely clustered patterns.}
\label{fig:tsne_M}

\end{figure}
%%%%%%%%%%%%%%%%%%%%%%%%%%%%%%%%%%%%%%%%%%%%%%%%%%%%%%%%%%%%5
To further analyze the performance of Audio-Visual Instruction design, we visualize the intermediate speech features that are contrastively aligned using Q-Former.
In particular, as discussed in Sec.~\ref{first_stage}, the contrastive audio-visual instruction alignment aims to extract audio embeddings closely relevant to the visual instructions. Consequently, the resulting audio embeddings are expected to include rich speaker state information.

As shown in Figure~\ref{fig:tsne_M}, we present samples of utterances representing five typical emotions. 
Notably, there is a discernible clustering pattern observed among embeddings associated with the same emotional type.
It is interesting that speech features belonging to the happiness class exhibit particularly close clustering, which could be attributed to the distinct characteristics of a happy voice.

\subsubsection{Generation Diversity of Talking Face Instruction System}
\begin{figure}[t]
    \centering
    \includegraphics[width=1.0\linewidth]{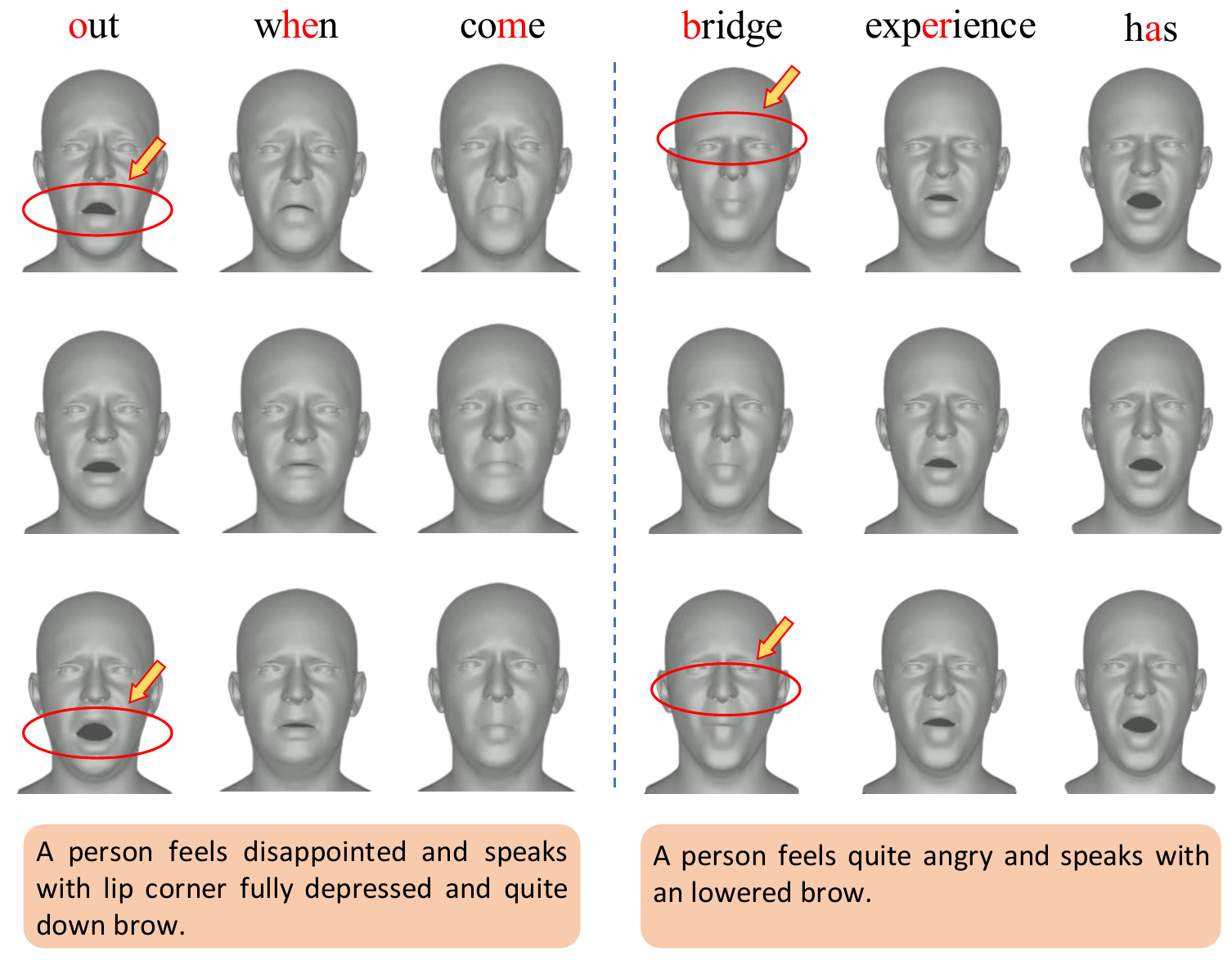}
    % \vspace{-10pt}
    \caption{Diverse generation results of the talking face instruction system are depicted. The bottom row showcases the audio-visual instruction, while the rows above demonstrate generation variations using the same text instruction. The left columns display the sad speaker status, where different lip curves are predicted, while the right columns demonstrate an angry case with varying eyebrow and cheek movements.
}
% \vspace{-12pt}
\label{fig:diversity}

\end{figure}

In Table~\ref{table:ablation_talking}, we illustrate the pivotal role of diffusion strategy in enhancing generation diversity. Additionally, in Figure~\ref{fig:diversity}, we present visualizations showcasing diverse synthesis. Observing the left column, it shows that multiple lip curves can be synthesized for instructions conveying disappointing emotions. Similarly, the right column demonstrates varied eyebrow and cheek movements in response to text instructions suggesting anger. These outcomes validate the capability of the talking face synthesis system to produce diverse results.

%%%%%%%%%%%%%%%%%%%%%%%%%%%%%%%%%%%%%%%%%%%%%%%%%%%%%%%%%%%%5
%%%%%%%%%%%%%%%%%%%%%%%%%%%%%%%%%%%%%%%%%%%%%%%%%%%%%%%%%%%%5

\subsubsection{OOD Analysis of Talking Face Instruction System}
\begin{figure}[t]
    \centering
    \includegraphics[width=1.0\linewidth]{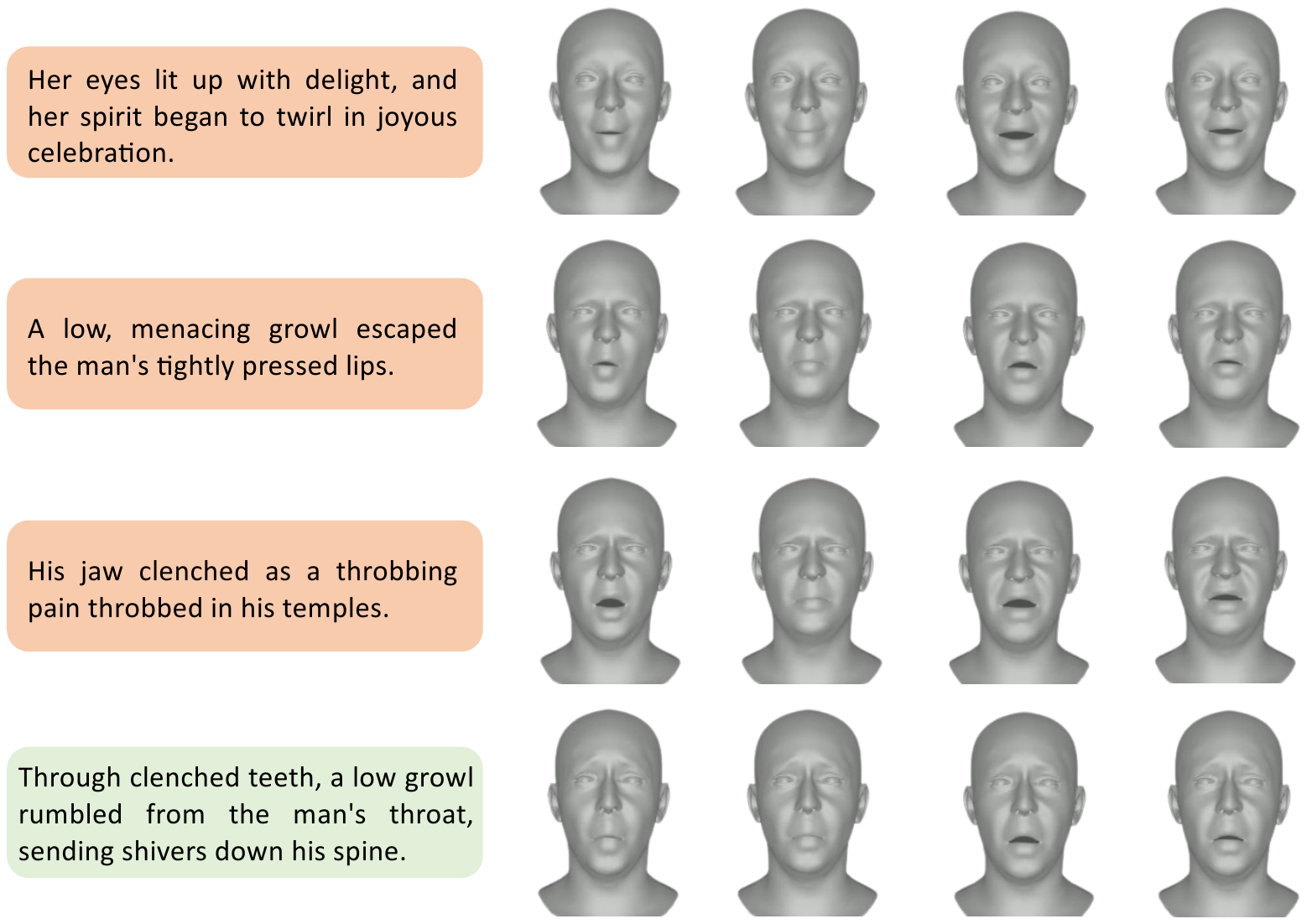}
    % \vspace{-10pt}
    \caption{Visualization of Out-of-Distribution (OOD) results from the Talking Face Instruction System. Within each row, we present instructed synthesis outcomes for the same speaker's speech, encompassing four distinct out-of-distribution instructions. The initial three rows showcase various successful cases while the final row illustrates an instance where the model misinterprets the instruction.
}
\vspace{-12pt}
\label{fig:ood}

\end{figure}

To further assess the generalizability of our proposed talking face synthesis module, we conducted experiments with out-of-distribution (OOD) instructions. Unlike the instructions in our dataset, which explicitly describe facial movements, we also explored visual instructions indicated by abstract concepts. As shown in Figure~\ref{fig:ood}, our model demonstrates the ability to capture the implicit speaking state of the speaker in the first three rows, yielding plausible synthesis results. This success can be attributed to the adoption of the diffusion mechanism and the structural similarity of natural language embeddings. However, when faced with particularly complex and abstract instructions, our model tends to misinterpret the implied speaking states as seen in the last row.

%%%%%%%%%%%%%%%%%%%%%%%%%%%%%%%%%%%%%%%%%%%%%%%%%%%%%%%%%%%%5
%%%%%%%%%%%%%%%%%%%%%%%%%%%%%%%%%%%%%%%%%%%%%%%%%%%%%%%%%%%%5

%% file: sections/conclusion.tex
\section{Conclusion}
\label{sec:conclusion}

In this paper, we propose \textbf{AVI-Talking}, an \textbf{Audio-Visual Instruction} system for expressive 3D \textbf{Talking} face generation.
We emphasize several appealing properties of our framework: 
1) We address the speech-driven expressive talking face generation by introducing an intermediate visual instruction, which decomposes the challenging audio-to-visual generation into two stages with clear learning objective. 
2) A soft prompting strategy is derived to harness the prior contextual knowledge underlying LLMs for speaker talking state comprehension. 
3) The disentangled talking prior learning procedure ensures complementary integration of lip-sync movements and audio-visual instruction. 
4) A diffusion prior network is introduced to map audio-visual instructions to latent distribution of content irrelevant space.

\textbf{Limitations}. Our model is currently trained on a labeled audio-visual instruction dataset. 1) We observe it exhibits insensitivity to certain specific speaking statuses. This phenomenon could be attributed to uneven data distribution, where certain speaking states are not adequately represented in the training dataset, making them challenging to discern from the speaker's speech.
2) The capabilities of the talking face synthesis network are limited to handling visual instructions closely aligned with the overall dataset distributions. As a consequence, for optimal instruction-following performance, users must provide instructions that closely resemble the predefined instructions.

\textbf{Future Work.} In this paper, we have investigated into specifying a pre-trained Large Language Model (LLM) for cross-modal audio-visual generation using finetuning techniques. 
Recent studies~\cite{10.5555/3495724.3496517} higlight the remarkable capability of Retrieval Augmented Generation (RAG) in injecting knowledge into Large Language Models (LLMs). Future research will involve comparing the effectiveness of RAG and fine-tuning performance, particularly tailored for this task.
Meanwhile, recent works~\cite{anonymous2024language} suggest that visual foundation models can yield competitive results, provided a robust visual tokenizer is utilized. Consequently, future research will delve into directly tokenizing stylized embeddings within the content-irrelevant space and fine-tuning general visual foundation models for expressive talking face synthesis. In this way, the model might be able to circumvent relying on specific audio-visual instruction dataset, thereby achieving superior performance with high generality.

% \textbf{Ethical Considerations.} Our method has the potential to be exploited for malicious purposes, such as generating deepfakes, which can have detrimental effects on various aspects of society, including misinformation and privacy breaches. To mitigate this risk and ensure responsible usage, we have decided to limit the licensing of our model strictly to research purposes and will share it exclusively with the deepfake detection community. In addition to licensing restrictions, we will proactively incorporate robust watermarks into the generation process to facilitate the identification and tracking of deepfakes generated using our method. These watermarks will serve as an essential tool for forensic analysis, thereby enhancing the resilience of our technology against potential misuse. 

%% file: main.bbl
\begin{thebibliography}{10}\itemsep=-1pt

\bibitem{alayrac2022flamingo}
Jean-Baptiste Alayrac, Jeff Donahue, Pauline Luc, Antoine Miech, Iain Barr, Yana Hasson, Karel Lenc, Arthur Mensch, Katherine Millican, Malcolm Reynolds, et~al.
\newblock Flamingo: a visual language model for few-shot learning.
\newblock {\em Advances in Neural Information Processing Systems}, 35:23716--23736, 2022.

\bibitem{anderson2016spice}
Peter Anderson, Basura Fernando, Mark Johnson, and Stephen Gould.
\newblock Spice: Semantic propositional image caption evaluation.
\newblock In {\em Computer Vision--ECCV 2016: 14th European Conference, Amsterdam, The Netherlands, October 11-14, 2016, Proceedings, Part V 14}, pages 382--398. Springer, 2016.

\bibitem{aneja2023facetalk}
Shivangi Aneja, Justus Thies, Angela Dai, and Matthias Nie{\ss}ner.
\newblock Facetalk: Audio-driven motion diffusion for neural parametric head models.
\newblock {\em arXiv preprint arXiv:2312.08459}, 2023.

\bibitem{anonymous2024language}
Anonymous.
\newblock Language model beats diffusion - tokenizer is key to visual generation.
\newblock In {\em The Twelfth International Conference on Learning Representations}, 2024.

\bibitem{baevski2020wav2vec}
Alexei Baevski, Yuhao Zhou, Abdelrahman Mohamed, and Michael Auli.
\newblock wav2vec 2.0: A framework for self-supervised learning of speech representations.
\newblock {\em Advances in neural information processing systems}, 33:12449--12460, 2020.

\bibitem{banerjee2005meteor}
Satanjeev Banerjee and Alon Lavie.
\newblock Meteor: An automatic metric for mt evaluation with improved correlation with human judgments.
\newblock In {\em Proceedings of the acl workshop on intrinsic and extrinsic evaluation measures for machine translation and/or summarization}, pages 65--72, 2005.

\bibitem{chen2020talking}
Lele Chen, Guofeng Cui, Celong Liu, Zhong Li, Ziyi Kou, Yi Xu, and Chenliang Xu.
\newblock Talking-head generation with rhythmic head motion.
\newblock In {\em European Conference on Computer Vision}, pages 35--51. Springer, 2020.

\bibitem{cudeiro2019capture}
Daniel Cudeiro, Timo Bolkart, Cassidy Laidlaw, Anurag Ranjan, and Michael~J Black.
\newblock Capture, learning, and synthesis of 3d speaking styles.
\newblock In {\em Proceedings of the IEEE/CVF Conference on Computer Vision and Pattern Recognition}, pages 10101--10111, 2019.

\bibitem{danvevcek2022emoca}
Radek Dan{\v{e}}{\v{c}}ek, Michael~J Black, and Timo Bolkart.
\newblock Emoca: Emotion driven monocular face capture and animation.
\newblock In {\em Proceedings of the IEEE/CVF Conference on Computer Vision and Pattern Recognition}, pages 20311--20322, 2022.

\bibitem{danvevcek2023emotional}
Radek Dan{\v{e}}{\v{c}}ek, Kiran Chhatre, Shashank Tripathi, Yandong Wen, Michael Black, and Timo Bolkart.
\newblock Emotional speech-driven animation with content-emotion disentanglement.
\newblock In {\em SIGGRAPH Asia 2023 Conference Papers}, pages 1--13, 2023.

\bibitem{esser2020taming}
Patrick Esser, Robin Rombach, and Björn Ommer.
\newblock Taming transformers for high-resolution image synthesis, 2020.

\bibitem{fan2022faceformer}
Yingruo Fan, Zhaojiang Lin, Jun Saito, Wenping Wang, and Taku Komura.
\newblock Faceformer: Speech-driven 3d facial animation with transformers.
\newblock In {\em Proceedings of the IEEE/CVF Conference on Computer Vision and Pattern Recognition}, pages 18770--18780, 2022.

\bibitem{gan2023efficient}
Yuan Gan, Zongxin Yang, Xihang Yue, Lingyun Sun, and Yi Yang.
\newblock Efficient emotional adaptation for audio-driven talking-head generation.
\newblock In {\em Proceedings of the IEEE/CVF International Conference on Computer Vision}, pages 22634--22645, 2023.

\bibitem{guo2021ad}
Yudong Guo, Keyu Chen, Sen Liang, Yong-Jin Liu, Hujun Bao, and Juyong Zhang.
\newblock Ad-nerf: Audio driven neural radiance fields for talking head synthesis.
\newblock In {\em Proceedings of the IEEE/CVF International Conference on Computer Vision}, pages 5784--5794, 2021.

\bibitem{gururani2023space}
Siddharth Gururani, Arun Mallya, Ting-Chun Wang, Rafael Valle, and Ming-Yu Liu.
\newblock Space: Speech-driven portrait animation with controllable expression.
\newblock In {\em Proceedings of the IEEE/CVF International Conference on Computer Vision}, pages 20914--20923, 2023.

\bibitem{heusel2017gans}
Martin Heusel, Hubert Ramsauer, Thomas Unterthiner, Bernhard Nessler, and Sepp Hochreiter.
\newblock Gans trained by a two time-scale update rule converge to a local nash equilibrium.
\newblock {\em Advances in neural information processing systems}, 30, 2017.

\bibitem{ho2020denoising}
Jonathan Ho, Ajay Jain, and Pieter Abbeel.
\newblock Denoising diffusion probabilistic models.
\newblock {\em Advances in neural information processing systems}, 33:6840--6851, 2020.

\bibitem{hsu2021hubert}
Wei-Ning Hsu, Benjamin Bolte, Yao-Hung~Hubert Tsai, Kushal Lakhotia, Ruslan Salakhutdinov, and Abdelrahman Mohamed.
\newblock Hubert: Self-supervised speech representation learning by masked prediction of hidden units.
\newblock {\em IEEE/ACM Transactions on Audio, Speech, and Language Processing}, 29:3451--3460, 2021.

\bibitem{huang2022towards}
Jie Huang and Kevin Chen-Chuan Chang.
\newblock Towards reasoning in large language models: A survey.
\newblock {\em arXiv preprint arXiv:2212.10403}, 2022.

\bibitem{huang2023audiogpt}
Rongjie Huang, Mingze Li, Dongchao Yang, Jiatong Shi, Xuankai Chang, Zhenhui Ye, Yuning Wu, Zhiqing Hong, Jiawei Huang, Jinglin Liu, et~al.
\newblock Audiogpt: Understanding and generating speech, music, sound, and talking head.
\newblock {\em arXiv preprint arXiv:2304.12995}, 2023.

\bibitem{ji2022eamm}
Xinya Ji, Hang Zhou, Kaisiyuan Wang, Qianyi Wu, Wayne Wu, Feng Xu, and Xun Cao.
\newblock Eamm: One-shot emotional talking face via audio-based emotion-aware motion model.
\newblock In {\em ACM SIGGRAPH 2022 Conference Proceedings}, pages 1--10, 2022.

\bibitem{ji2021audio}
Xinya Ji, Hang Zhou, Kaisiyuan Wang, Wayne Wu, Chen~Change Loy, Xun Cao, and Feng Xu.
\newblock Audio-driven emotional video portraits.
\newblock In {\em Proceedings of the IEEE/CVF conference on computer vision and pattern recognition}, pages 14080--14089, 2021.

\bibitem{karras2017audio}
Tero Karras, Timo Aila, Samuli Laine, Antti Herva, and Jaakko Lehtinen.
\newblock Audio-driven facial animation by joint end-to-end learning of pose and emotion.
\newblock {\em ACM Transactions on Graphics (TOG)}, 36(4):1--12, 2017.

\bibitem{lam2022bddm}
Max W.~Y. Lam, Jun Wang, Dan Su, and Dong Yu.
\newblock {BDDM}: Bilateral denoising diffusion models for fast and high-quality speech synthesis.
\newblock In {\em International Conference on Learning Representations}, 2022.

\bibitem{10.5555/3495724.3496517}
Patrick Lewis, Ethan Perez, Aleksandra Piktus, Fabio Petroni, Vladimir Karpukhin, Naman Goyal, Heinrich K\"{u}ttler, Mike Lewis, Wen-tau Yih, Tim Rockt\"{a}schel, Sebastian Riedel, and Douwe Kiela.
\newblock Retrieval-augmented generation for knowledge-intensive nlp tasks.
\newblock In {\em Proceedings of the 34th International Conference on Neural Information Processing Systems}, NIPS'20, Red Hook, NY, USA, 2020. Curran Associates Inc.

\bibitem{li2023blip}
Junnan Li, Dongxu Li, Silvio Savarese, and Steven Hoi.
\newblock Blip-2: Bootstrapping language-image pre-training with frozen image encoders and large language models.
\newblock {\em arXiv preprint arXiv:2301.12597}, 2023.

\bibitem{FLAME:SiggraphAsia2017}
Tianye Li, Timo Bolkart, Michael.~J. Black, Hao Li, and Javier Romero.
\newblock Learning a model of facial shape and expression from {4D} scans.
\newblock {\em ACM Transactions on Graphics, (Proc. SIGGRAPH Asia)}, 36(6):194:1--194:17, 2017.

\bibitem{liang2022expressive}
Borong Liang, Yan Pan, Zhizhi Guo, Hang Zhou, Zhibin Hong, Xiaoguang Han, Junyu Han, Jingtuo Liu, Errui Ding, and Jingdong Wang.
\newblock Expressive talking head generation with granular audio-visual control.
\newblock In {\em Proceedings of the IEEE/CVF Conference on Computer Vision and Pattern Recognition}, pages 3387--3396, 2022.

\bibitem{liang2022mind}
Victor~Weixin Liang, Yuhui Zhang, Yongchan Kwon, Serena Yeung, and James~Y Zou.
\newblock Mind the gap: Understanding the modality gap in multi-modal contrastive representation learning.
\newblock {\em Advances in Neural Information Processing Systems}, 35:17612--17625, 2022.

\bibitem{lin2004rouge}
Chin-Yew Lin.
\newblock Rouge: A package for automatic evaluation of summaries.
\newblock In {\em Text summarization branches out}, pages 74--81, 2004.

\bibitem{livingstone2018ryerson}
Steven~R Livingstone and Frank~A Russo.
\newblock The ryerson audio-visual database of emotional speech and song (ravdess): A dynamic, multimodal set of facial and vocal expressions in north american english.
\newblock {\em PloS one}, 13(5):e0196391, 2018.

\bibitem{ma2023talkclip}
Yifeng Ma, Suzhen Wang, Yu Ding, Bowen Ma, Tangjie Lv, Changjie Fan, Zhipeng Hu, Zhidong Deng, and Xin Yu.
\newblock Talkclip: Talking head generation with text-guided expressive speaking styles.
\newblock {\em arXiv preprint arXiv:2304.00334}, 2023.

\bibitem{ma2023styletalk}
Yifeng Ma, Suzhen Wang, Zhipeng Hu, Changjie Fan, Tangjie Lv, Yu Ding, Zhidong Deng, and Xin Yu.
\newblock Styletalk: One-shot talking head generation with controllable speaking styles.
\newblock {\em arXiv preprint arXiv:2301.01081}, 2023.

\bibitem{ma2023dreamtalk}
Yifeng Ma, Shiwei Zhang, Jiayu Wang, Xiang Wang, Yingya Zhang, and Zhidong Deng.
\newblock Dreamtalk: When expressive talking head generation meets diffusion probabilistic models.
\newblock {\em arXiv preprint arXiv:2312.09767}, 2023.

\bibitem{mohamed2021arabic}
Omar Mohamed and Salah~A Aly.
\newblock Arabic speech emotion recognition employing wav2vec2. 0 and hubert based on baved dataset.
\newblock {\em arXiv preprint arXiv:2110.04425}, 2021.

\bibitem{ng2023can}
Evonne Ng, Sanjay Subramanian, Dan Klein, Angjoo Kanazawa, Trevor Darrell, and Shiry Ginosar.
\newblock Can language models learn to listen?
\newblock In {\em Proceedings of the IEEE/CVF International Conference on Computer Vision}, pages 10083--10093, 2023.

\bibitem{oord2018representation}
Aaron van~den Oord, Yazhe Li, and Oriol Vinyals.
\newblock Representation learning with contrastive predictive coding.
\newblock {\em arXiv preprint arXiv:1807.03748}, 2018.

\bibitem{NEURIPS2022_b1efde53}
Long Ouyang, Jeffrey Wu, Xu Jiang, Diogo Almeida, Carroll Wainwright, Pamela Mishkin, Chong Zhang, Sandhini Agarwal, Katarina Slama, Alex Ray, John Schulman, Jacob Hilton, Fraser Kelton, Luke Miller, Maddie Simens, Amanda Askell, Peter Welinder, Paul~F Christiano, Jan Leike, and Ryan Lowe.
\newblock Training language models to follow instructions with human feedback.
\newblock In S. Koyejo, S. Mohamed, A. Agarwal, D. Belgrave, K. Cho, and A. Oh, editors, {\em Advances in Neural Information Processing Systems}, volume~35, pages 27730--27744. Curran Associates, Inc., 2022.

\bibitem{papineni2002bleu}
Kishore Papineni, Salim Roukos, Todd Ward, and Wei-Jing Zhu.
\newblock Bleu: a method for automatic evaluation of machine translation.
\newblock In {\em Proceedings of the 40th annual meeting of the Association for Computational Linguistics}, pages 311--318, 2002.

\bibitem{paszke2019pytorch}
Adam Paszke, Sam Gross, Francisco Massa, Adam Lerer, James Bradbury, Gregory Chanan, Trevor Killeen, Zeming Lin, Natalia Gimelshein, Luca Antiga, et~al.
\newblock Pytorch: An imperative style, high-performance deep learning library.
\newblock {\em Advances in neural information processing systems}, 32, 2019.

\bibitem{peng2023emotalk}
Ziqiao Peng, Haoyu Wu, Zhenbo Song, Hao Xu, Xiangyu Zhu, Jun He, Hongyan Liu, and Zhaoxin Fan.
\newblock Emotalk: Speech-driven emotional disentanglement for 3d face animation.
\newblock In {\em Proceedings of the IEEE/CVF International Conference on Computer Vision}, pages 20687--20697, 2023.

\bibitem{prajwal2020lip}
KR Prajwal, Rudrabha Mukhopadhyay, Vinay~P Namboodiri, and CV Jawahar.
\newblock A lip sync expert is all you need for speech to lip generation in the wild.
\newblock In {\em Proceedings of the 28th ACM international conference on multimedia}, pages 484--492, 2020.

\bibitem{ramesh2022hierarchical}
Aditya Ramesh, Prafulla Dhariwal, Alex Nichol, Casey Chu, and Mark Chen.
\newblock Hierarchical text-conditional image generation with clip latents.
\newblock {\em arXiv preprint arXiv:2204.06125}, 1(2):3, 2022.

\bibitem{ren2023diffusion}
Zhiyuan Ren, Zhihong Pan, Xin Zhou, and Le Kang.
\newblock Diffusion motion: Generate text-guided 3d human motion by diffusion model.
\newblock In {\em ICASSP 2023-2023 IEEE International Conference on Acoustics, Speech and Signal Processing (ICASSP)}, pages 1--5. IEEE, 2023.

\bibitem{richard2021audio}
Alexander Richard, Colin Lea, Shugao Ma, Jurgen Gall, Fernando De~la Torre, and Yaser Sheikh.
\newblock Audio-and gaze-driven facial animation of codec avatars.
\newblock In {\em Proceedings of the IEEE/CVF winter conference on applications of computer vision}, pages 41--50, 2021.

\bibitem{richard2021meshtalk}
Alexander Richard, Michael Zollh{\"o}fer, Yandong Wen, Fernando De~la Torre, and Yaser Sheikh.
\newblock Meshtalk: 3d face animation from speech using cross-modality disentanglement.
\newblock In {\em Proceedings of the IEEE/CVF International Conference on Computer Vision}, pages 1173--1182, 2021.

\bibitem{sadoughi2019speech}
Najmeh Sadoughi and Carlos Busso.
\newblock Speech-driven expressive talking lips with conditional sequential generative adversarial networks.
\newblock {\em IEEE Transactions on Affective Computing}, 12(4):1031--1044, 2019.

\bibitem{schick2023toolformer}
Timo Schick, Jane Dwivedi-Yu, Roberto Dess{\`\i}, Roberta Raileanu, Maria Lomeli, Luke Zettlemoyer, Nicola Cancedda, and Thomas Scialom.
\newblock Toolformer: Language models can teach themselves to use tools.
\newblock {\em arXiv preprint arXiv:2302.04761}, 2023.

\bibitem{sinha2022emotion}
Sanjana Sinha, Sandika Biswas, Ravindra Yadav, and Brojeshwar Bhowmick.
\newblock Emotion-controllable generalized talking face generation.
\newblock {\em arXiv preprint arXiv:2205.01155}, 2022.

\bibitem{sun2023imagebrush}
Yasheng Sun, Yifan Yang, Houwen Peng, Yifei Shen, Yuqing Yang, Han Hu, Lili Qiu, and Hideki Koike.
\newblock Imagebrush: Learning visual in-context instructions for exemplar-based image manipulation.
\newblock {\em arXiv preprint arXiv:2308.00906}, 2023.

\bibitem{sun2021speech2talking}
Yasheng Sun, Hang Zhou, Ziwei Liu, and Hideki Koike.
\newblock Speech2talking-face: Inferring and driving a face with synchronized audio-visual representation.
\newblock In {\em IJCAI}, volume~2, page~4, 2021.

\bibitem{sun2022masked}
Yasheng Sun, Hang Zhou, Kaisiyuan Wang, Qianyi Wu, Zhibin Hong, Jingtuo Liu, Errui Ding, Jingdong Wang, Ziwei Liu, and Koike Hideki.
\newblock Masked lip-sync prediction by audio-visual contextual exploitation in transformers.
\newblock In {\em SIGGRAPH Asia 2022 Conference Papers}, pages 1--9, 2022.

\bibitem{sun2023diffposetalk}
Zhiyao Sun, Tian Lv, Sheng Ye, Matthieu~Gaetan Lin, Jenny Sheng, Yu-Hui Wen, Minjing Yu, and Yong-jin Liu.
\newblock Diffposetalk: Speech-driven stylistic 3d facial animation and head pose generation via diffusion models.
\newblock {\em arXiv preprint arXiv:2310.00434}, 2023.

\bibitem{tan2023emmn}
Shuai Tan, Bin Ji, and Ye Pan.
\newblock Emmn: Emotional motion memory network for audio-driven emotional talking face generation.
\newblock In {\em Proceedings of the IEEE/CVF International Conference on Computer Vision}, pages 22146--22156, 2023.

\bibitem{tevet2022human}
Guy Tevet, Sigal Raab, Brian Gordon, Yonatan Shafir, Daniel Cohen-Or, and Amit~H Bermano.
\newblock Human motion diffusion model.
\newblock {\em arXiv preprint arXiv:2209.14916}, 2022.

\bibitem{thambiraja2023imitator}
Balamurugan Thambiraja, Ikhsanul Habibie, Sadegh Aliakbarian, Darren Cosker, Christian Theobalt, and Justus Thies.
\newblock Imitator: Personalized speech-driven 3d facial animation.
\newblock In {\em Proceedings of the IEEE/CVF International Conference on Computer Vision}, pages 20621--20631, 2023.

\bibitem{thies2020neural}
Justus Thies, Mohamed Elgharib, Ayush Tewari, Christian Theobalt, and Matthias Nie{\ss}ner.
\newblock Neural voice puppetry: Audio-driven facial reenactment.
\newblock In {\em Computer Vision--ECCV 2020: 16th European Conference, Glasgow, UK, August 23--28, 2020, Proceedings, Part XVI 16}, pages 716--731. Springer, 2020.

\bibitem{Touvron2023LLaMAOA}
Hugo Touvron, Thibaut Lavril, Gautier Izacard, Xavier Martinet, Marie-Anne Lachaux, Timoth{\'e}e Lacroix, Baptiste Rozi{\`e}re, Naman Goyal, Eric Hambro, Faisal Azhar, Aurelien Rodriguez, Armand Joulin, Edouard Grave, and Guillaume Lample.
\newblock Llama: Open and efficient foundation language models.
\newblock {\em ArXiv}, abs/2302.13971, 2023.

\bibitem{van2016conditional}
Aaron Van~den Oord, Nal Kalchbrenner, Lasse Espeholt, Oriol Vinyals, Alex Graves, et~al.
\newblock Conditional image generation with pixelcnn decoders.
\newblock {\em Advances in neural information processing systems}, 29, 2016.

\bibitem{vougioukas2020realistic}
Konstantinos Vougioukas, Stavros Petridis, and Maja Pantic.
\newblock Realistic speech-driven facial animation with gans.
\newblock {\em International Journal of Computer Vision}, 128:1398--1413, 2020.

\bibitem{wang2023agentavatar}
Duomin Wang, Bin Dai, Yu Deng, and Baoyuan Wang.
\newblock Agentavatar: Disentangling planning, driving and rendering for photorealistic avatar agents.
\newblock {\em arXiv preprint arXiv:2311.17465}, 2023.

\bibitem{wang2020mead}
Kaisiyuan Wang, Qianyi Wu, Linsen Song, Zhuoqian Yang, Wayne Wu, Chen Qian, Ran He, Yu Qiao, and Chen~Change Loy.
\newblock Mead: A large-scale audio-visual dataset for emotional talking-face generation.
\newblock In {\em European Conference on Computer Vision}, pages 700--717. Springer, 2020.

\bibitem{kaisiyuan2020mead}
Kaisiyuan Wang, Qianyi Wu, Linsen Song, Zhuoqian Yang, Wayne Wu, Chen Qian, Ran He, Yu Qiao, and Chen~Change Loy.
\newblock Mead: A large-scale audio-visual dataset for emotional talking-face generation.
\newblock In {\em ECCV}, 2020.

\bibitem{wang2023survey}
Lei Wang, Chen Ma, Xueyang Feng, Zeyu Zhang, Hao Yang, Jingsen Zhang, Zhiyuan Chen, Jiakai Tang, Xu Chen, Yankai Lin, et~al.
\newblock A survey on large language model based autonomous agents.
\newblock {\em arXiv preprint arXiv:2308.11432}, 2023.

\bibitem{wang2019describing}
Qingzhong Wang and Antoni~B Chan.
\newblock Describing like humans: on diversity in image captioning.
\newblock In {\em Proceedings of the IEEE/CVF Conference on Computer Vision and Pattern Recognition}, pages 4195--4203, 2019.

\bibitem{wei2022emergent}
Jason Wei, Yi Tay, Rishi Bommasani, Colin Raffel, Barret Zoph, Sebastian Borgeaud, Dani Yogatama, Maarten Bosma, Denny Zhou, Donald Metzler, et~al.
\newblock Emergent abilities of large language models.
\newblock {\em arXiv preprint arXiv:2206.07682}, 2022.

\bibitem{wei2022chain}
Jason Wei, Xuezhi Wang, Dale Schuurmans, Maarten Bosma, Fei Xia, Ed Chi, Quoc~V Le, Denny Zhou, et~al.
\newblock Chain-of-thought prompting elicits reasoning in large language models.
\newblock {\em Advances in Neural Information Processing Systems}, 35:24824--24837, 2022.

\bibitem{wiles2018x2face}
Olivia Wiles, A Koepke, and Andrew Zisserman.
\newblock X2face: A network for controlling face generation using images, audio, and pose codes.
\newblock In {\em Proceedings of the European conference on computer vision (ECCV)}, pages 670--686, 2018.

\bibitem{wu2021imitating}
Haozhe Wu, Jia Jia, Haoyu Wang, Yishun Dou, Chao Duan, and Qingshan Deng.
\newblock Imitating arbitrary talking style for realistic audio-driven talking face synthesis.
\newblock In {\em Proceedings of the 29th ACM International Conference on Multimedia}, pages 1478--1486, 2021.

\bibitem{wu2023nextgpt}
Shengqiong Wu, Hao Fei, Leigang Qu, Wei Ji, and Tat-Seng Chua.
\newblock Next-gpt: Any-to-any multimodal llm, 2023.

\bibitem{xing2023codetalker}
Jinbo Xing, Menghan Xia, Yuechen Zhang, Xiaodong Cun, Jue Wang, and Tien-Tsin Wong.
\newblock Codetalker: Speech-driven 3d facial animation with discrete motion prior.
\newblock In {\em Proceedings of the IEEE/CVF Conference on Computer Vision and Pattern Recognition}, pages 12780--12790, 2023.

\bibitem{xu2023high}
Chao Xu, Junwei Zhu, Jiangning Zhang, Yue Han, Wenqing Chu, Ying Tai, Chengjie Wang, Zhifeng Xie, and Yong Liu.
\newblock High-fidelity generalized emotional talking face generation with multi-modal emotion space learning.
\newblock In {\em Proceedings of the IEEE/CVF Conference on Computer Vision and Pattern Recognition}, pages 6609--6619, 2023.

\bibitem{xu2023secap}
Yaoxun Xu, Hangting Chen, Jianwei Yu, Qiaochu Huang, Zhiyong Wu, Shixiong Zhang, Guangzhi Li, Yi Luo, and Rongzhi Gu.
\newblock Secap: Speech emotion captioning with large language model.
\newblock {\em arXiv preprint arXiv:2312.10381}, 2023.

\bibitem{yu2023talking}
Zhentao Yu, Zixin Yin, Deyu Zhou, Duomin Wang, Finn Wong, and Baoyuan Wang.
\newblock Talking head generation with probabilistic audio-to-visual diffusion priors.
\newblock In {\em Proceedings of the IEEE/CVF International Conference on Computer Vision}, pages 7645--7655, 2023.

\bibitem{zeng2023glmb}
Aohan Zeng, Xiao Liu, Zhengxiao Du, Zihan Wang, Hanyu Lai, Ming Ding, Zhuoyi Yang, Yifan Xu, Wendi Zheng, Xiao Xia, Weng~Lam Tam, Zixuan Ma, Yufei Xue, Jidong Zhai, Wenguang Chen, Zhiyuan Liu, Peng Zhang, Yuxiao Dong, and Jie Tang.
\newblock {GLM}-130b: An open bilingual pre-trained model.
\newblock In {\em The Eleventh International Conference on Learning Representations}, 2023.

\bibitem{Zhang2022OPTOP}
Susan Zhang, Stephen Roller, Naman Goyal, Mikel Artetxe, Moya Chen, Shuohui Chen, Christopher Dewan, Mona~T. Diab, Xian Li, Xi~Victoria Lin, Todor Mihaylov, Myle Ott, Sam Shleifer, Kurt Shuster, Daniel Simig, Punit~Singh Koura, Anjali Sridhar, Tianlu Wang, and Luke Zettlemoyer.
\newblock Opt: Open pre-trained transformer language models.
\newblock {\em ArXiv}, abs/2205.01068, 2022.

\bibitem{zhang2023sadtalker}
Wenxuan Zhang, Xiaodong Cun, Xuan Wang, Yong Zhang, Xi Shen, Yu Guo, Ying Shan, and Fei Wang.
\newblock Sadtalker: Learning realistic 3d motion coefficients for stylized audio-driven single image talking face animation.
\newblock In {\em Proceedings of the IEEE/CVF Conference on Computer Vision and Pattern Recognition}, pages 8652--8661, 2023.

\bibitem{zheng2023minigpt5}
Kaizhi Zheng, Xuehai He, and Xin~Eric Wang.
\newblock Minigpt-5: Interleaved vision-and-language generation via generative vokens, 2023.

\bibitem{zhou2021pose}
Hang Zhou, Yasheng Sun, Wayne Wu, Chen~Change Loy, Xiaogang Wang, and Ziwei Liu.
\newblock Pose-controllable talking face generation by implicitly modularized audio-visual representation.
\newblock In {\em Proceedings of the IEEE/CVF conference on computer vision and pattern recognition}, pages 4176--4186, 2021.

\bibitem{zhu2023minigpt}
Deyao Zhu, Jun Chen, Xiaoqian Shen, Xiang Li, and Mohamed Elhoseiny.
\newblock Minigpt-4: Enhancing vision-language understanding with advanced large language models.
\newblock {\em arXiv preprint arXiv:2304.10592}, 2023.

\end{thebibliography}
